# The RSNA Intracranial Aneurysm (RSNA-ICA) Dataset

Maria Correia de Verdier[1], Rachit Saluja[2,3], Jason Sho[4], Maryam Vabarizad[4], Rennie Yung-Chieh Chen[5], Uyen N. T. Nguyen[6], Mona Alrehaili[7], Layal Aweidah[8], Deniz Bulja[9], Wesley C. Chan[10], Hernan Chaves[11], Madhavi Duvvuri[12], Huseyin Ekin Ergin[13], Undrakh-Erdene Erdenebold[14], Ekim Gumeler[15], Mohamed Sobhi Jabal[16], Chin-Chi Kuo[17,18,19,20,21], Fatima Mubarak[22], Sevde Nur Emir[23], Scott Riley K. Ong[24], Johanna Ortiz[25], Almudena Pérez-Lara[26], Andreas M. Rauschecker[12], Shayan Sirat Maheen Anwar[22], Charit Tippareddy[27], Tam Tran[28], Sorawis Visrutaratna[29], John Mongan[30], Adam E. Flanders[31], Robyn Ball[32], Greg Zaharchuk[33], Peter D. Chang[34], Felipe Kitamura[35], Errol Colak[36], Luciano Prevedello[37], Tyler Richards[38], Evan Calabrese[16*], Jeffrey D. Rudie[39,40*]

**Evan Calabrese and Jeffrey D. Rudie contributed equally as co-senior authors*

[1]Department of Surgical Sciences, Section of Neuroradiology, Uppsala University, Uppsala, Sweden, [2]Department of Electrical and Computer Engineering, Cornell University and Cornell Tech, New York, NY, USA, [3]Department of Radiology, Weill Cornell Medicine, New York, NY, USA, [4]Informatics Department, Radiological Society of North America, Oak Brook, IL, USA, [5]Taipei Medical University Hospital, Taipei, Taiwan, [6]Department of Radiology, UC San Diego, San Diego, CA, USA, [7]Department of Radiology, Prince Sultan Military Medical City, Riyadh, Saudi Arabia, [8]Interventional Radiology, Liverpool Hospital, Sydney, Australia, [9]Clinical Center of the University of Sarajevo, Sarajevo, Bosnia and Herzegovina, [10]Health Sciences Centre, Memorial University of Newfoundland, Canada, [11]Departamento de Diagnóstico por Imágenes, Fleni, Buenos Aires, Argentina, [12]Center for Intelligent Imaging, Department of Radiology & Biomedical Imaging, University of California San Francisco, San Francisco, CA, USA, [13]Koç University School of Medicine, Istanbul, Turkey, [14]Department of Radiology, Intermed IUHW Hospital, International Medical Center, Ulaanbaatar, Mongolia, [15]Department of Radiology, Hacettepe University, Faculty of Medicine, Ankara, Turkey, [16]Department of Radiology, Duke University, Durham, NC, USA, [17]Big Data Center, China Medical University Hospital, Taichung, Taiwan, [18]Division of Nephrology, Department of Internal Medicine, China

Medical University Hospital, Taichung, Taiwan, [19]Department of Biomedical Informatics, School of Medicine, China Medical University, Taichung, Taiwan, [20]Department of Bioinformatics and Medical Engineering, Asia University, Taichung, Taiwan, [21]Office of Research and Development, Asia University, Taichung, Taiwan, [22]Aga Khan University Hospital, Pakistan, [23]University of Health Sciences, Ümraniye Training and Research Hospital, Istanbul, Turkey, [24]Department of Radiology, Philippine General Hospital, University of the Philippines Manila, Manila, Philippines, [25]Department of Radiology, Kingston Health Sciences Centre, Kingston, ON, Canada, [26]Radiology Department, Hospital Regional Universitario de Málaga, Málaga, Spain, [27]University Hospitals Cleveland Medical Center, Cleveland, OH, USA, [28]Gold Coast Hospital and Health Service, Queensland, Australia, [29]Department of Radiology, Faculty of Medicine, Chiang Mai University, Chiang Mai, Thailand, [30]Department of Radiology & Biomedical Imaging; Department of Medicine, Division of Clinical Informatics and Digital Transformation, University of California San Francisco, San Francisco, CA, USA, [31]Department of Radiology, Division of Neuroradiology/ENT, Thomas Jefferson University, Philadelphia, PA, USA, [32]The Jackson Laboratory, Bar Harbor, ME, USA, [33]Department of Radiology, Stanford University School of Medicine, Palo Alto, CA, USA, [34]University of California Irvine, Irvine, CA, USA, [35]Department of Diagnostic Imaging, Universidade Federal de São Paulo, São Paulo, Brazil, [36]Department of Medical Imaging, University of Toronto, Toronto, Canada, [37]Department of Radiology, Ohio State University, Columbus, OH, USA, [38]Department of Radiology and Imaging sciences, University of Utah School of medicine, Salt Lake City, UT, USA, [39]Department of Radiology, University of California San Diego, San Diego, CA, USA, [40]Department of Radiology, Scripps Clinic Medical Group, San Diego, CA, USA

## Data Contributor Group

Matt Aitken[1], Abdulrahman Alharbi[2], Odej Ali Abud MD[3], Salita Angkurawaranon MD[4], Bolorsaikhan Baatar MD, MSc[5], Clarisse Dianne B. Cayetano[6], Lin Che-Chen MS[7], Luke Christopher Smith[8], Ross Copping[9], Gonchigsuren Dagvasumberel MD., PhD[10], Munkhbaatar Dagvasumberel MD., Ph.D[11], Yunus Emre Senturk[12], Florencia Ferraudo[13], Hui-Chao Tsai MS[7], Memoona Khan[14], Ashmitha Kumar[1], Alexandra MacMaster MD[15], Justin Damian M. Malubay[6], Gowri Menon[1], Kruscica Nedim[16],

Shady Osman[9], Jacob J. Peoples PhD[17], Lejla Prnjavorac MD[3], Gemma Rojas Escobar[18], Mert R. Sabuncu[19], Glen Schlaphoff[9], Mina'a Shahid[14], Sheng-Hsuan Chen MS[7], Alex Shoung[1], Jennifer Sommer DO[20], Phattanun Thabarsa[4], Tanveer ul Haq[14], Muhammad Yousaf[14], Burhan Zafar[14,21]

[1]Gold Coast Hospital and Health Service, Queensland, Australia, [2]Department of Radiology, Prince Sultan Military Medical City, Riyadh, Saudi Arabia, [3]Clinical Center of the University of Sarajevo, Bosnia and Herzegovina, [4]Department of Radiology, Faculty of Medicine, Chiang Mai University, Chiang Mai, Thailand, [5]Mungunguur Hospital, Ulaanbaatar, Mongolia, [6]Department of Radiology, Philippine General Hospital, University of the Philippines Manila, Manila, Philippines, [7]Big Data Center, China Medical University Hospital, Taichung, Taiwan, [8]South Australia Medical Imaging, Central Adelaide Local Health Network, Adelaide, Australia, [9]Interventional Radiology, Liverpool Hospital, Sydney, Australia, [10]Grandmed Hospital, Ulaanbaatar, Mongolia, [11]Department of Radiology School of Medicine, Mongolian National University of Medical Sciences, Ulaanbaatar, Mongolia, [12]Koç University School of Medicine, Istanbul, Turkey, [13]Departamento de Diagnóstico por Imágenes, Fleni, Buenos Aires, Argentina, [14]Aga Khan University Hospital, Pakistan, [15]Queen's University, Kingston, ON, Canada, [16]EHS d.o.o., Bosnia and Herzegovina, [17]Department of Radiology, Memorial Sloan Kettering Cancer Center, New York, NY, USA, [18]Radiology Department, Hospital Regional Universitario de Málaga, Málaga, Spain, [19]Electrical and Computer Engineering, Cornell Tech, New York, NY, USA, [20]University Hospitals Cleveland Medical Center, Cleveland, OH, USA, [21]Department of Radiology, Weill Cornell Medicine, New York, NY, USA

## Dataset Annotator Group

Shehbaz Ansari[1], Leon Bos[2], M. Travis Caton[3], Jiawei Chen[4], Gagandeep Choudhary[5], Evangelia Christodoulou[6], Sammy Chu[7], Charlotte Yuk-Yan Chung[8], Marco Colasurdo[9], S. A. Cornelissen[10], Sean Creeden[11], Eduardo Portela de Oliveira[12], Jamal Derakhshan[13], Mircea C. Dobre[14], Siddhant Dogra[15], Venkata Dola[16], Anat Yahav Dovrat[17], Javed Khader Eliyas[18], Mohamed Elshikh[19], Augusto Lio da Mota Goncalves Filho[20], Colbey W. Freeman[21], Margalin Dragos Gabriel[22],

Malik Galijasevic[23], Vineet Vijay Gorolay[24,25], Arsany Hakim[26], Roula Hourani[27], Houjing Huang[28], Ali Hussain[29], John Huston[30], Michael Iv[31], Anthony Kam[32,33], Anthony Kam[34], Shreyas Reddy Kankara[35], Ryan T. Kellogg[36], Shahriar Kolahi[37], Marinos Kontzialis[38], Yan Lu[4], Weeratunge Mudiyanselage Nishantha Malalasekera[39], Franklin A. Marden[40], Jignesh Modi[41], Muthupalaniappaan Muthappan[42], Martin D. Ollenschleger[43], Julio Cesar Palomino Oquendo[44], David Ozretic[45], Markand Patel[46], Alida Postma[47], Ramya Shankaranarayana Rao[48], Pradeep Raj Regmi[49], Vivien Richter[50], Tanvir Rizvi[36], Alaa Saleh[36], Marianne Schell[51], Eric D. Schwartz[52], Robert Shih[53], Pengcheng Shi[4], Gagandeep Singh[54], Aapo Sirén[55], Juan E. Small[56], Neetu Soni[57], Houman Sotoudeh[58], Katie Suzanne Traylor[59], Benjamin Turner[60], Cristina Utrilla[61], Kaiyuan Yang[28], Ismail Yardimcioglu[62]

[1]Rush University Medical Center, Chicago, Illinois, USA, [2]Department of Radiology, NoordWest Ziekenhuisgroep, Alkmaar, The Netherlands, [3]Department of Neurosurgery, Mount Sinai Hospital, New York, New York, USA, [4]Medical Image Insights Co. Ltd., Shanghai, China, [5]Radiology, Oregon Health and Science University, Portland, Oregon, USA, [6]Department of Radiology, General University Hospital of Patras, Rion, Greece, [7]University of Washington, Seattle, Washington, USA, [8]Department of Neuroradiology, NYU Langone Health, New York, New York, USA, [9]Oregon Health and Science University, Portland, Oregon, USA, [10]Department of Radiology, Erasmus University Medical Center, Rotterdam, Zuid-Holland, The Netherlands, [11]UICOMP, Department of Neuroradiology, Peoria, Illinois, USA, [12]The Ottawa Hospital - University of Ottawa, Ottawa, Ontario, Canada, [13]Department of Radiology, University of Pennsylvania, Philadelphia, Pennsylvania, USA, [14]U.S. Dept of Veterans Affairs, Palo Alto, California, USA, [15]Department of Radiology, NYU Langone Health, New York, New York, USA, [16]George Washington University, Washington, DC, USA, [17]University of Toronto, Toronto, Ontario, Canada, [18]Neurosurgery, Saint Francis Hospital, Hartford, Connecticut, USA, [19]Department of Radiology, University of Iowa, Iowa City, Iowa, USA, [20]University of Ottawa, Ottawa, Ontario, Canada, [21]Department of Radiology, University of Pennsylvania, USA, [22]Spitalul Universitar de Urgenta Bucuresti, București, Romania, [23]University Clinic for Radiology - Medical University of Innsbruck, Innsbruck, Austria, [24]Department of

Radiology, Royal North Shore Hospital, Sydney, New South Wales, Australia, [25]Department of Radiology, Royal Prince Alfred Hospital, Sydney, New South Wales, Australia, [26]University Institute of Diagnostic and Interventional Neuroradiology, Bern University Hospital Inselspital, Bern, Switzerland, [27]American University of Beirut Medical Center, Beirut, Lebanon, [28]Department of Quantitative Biomedicine, University of Zurich, Zurich, Switzerland, [29]University of Rochester Medical Center, Rochester, New York, USA, [30]Department of Radiology, Mayo Clinic, Rochester, Minnesota, USA, [31]Department of Radiology, Stanford University, Palo Alto, California, USA, [32]Division of Neuroradiology, The Alfred, Melbourne, VIC, Australia, [33]Massachusetts General Hospital, Harvard Medical School, Boston, USA, [34]Departments of Radiology and Neurosurgery, Loyola University Medical Center, Maywood, Illinois, USA, [35]St John's Medical College Hospital, Bengaluru, Karnataka, India, [36]University of Virginia, Charlottesville, Virginia, USA, [37]Advanced Diagnostic and Interventional Radiology Research Center (ADIR), Radiology Department, Imam Khomeini Hospital, Tehran University of Medical Sciences, Tehran, Iran, [38]Department of Radiology, Division of Neuroradiology, Northwestern University, Chicago, Illinois, USA, [39]Wessex Neurological Centre - Southampton General Hospital, Southampton, Hampshire, United Kingdom, [40]Advocate Aurora Health System, Chicago, Illinois, USA, [41]Midstate Medical Center, Hartford Heathcare, Meriden, Connecticut, USA, [42]Wellington Regional Hospital, Wellington, New Zealand, [43]Department of Radiology, Hartford Hospital, Hartford, Connecticut, USA, [44]Department of Neurosurgery, Ribera POVISA Hospital, Vigo, Spain, [45]University Hospital Centre Zagreb, Zagreb, Croatia, [46]Imperial College Healthcare NHS Trust, London, United Kingdom, [47]Department of Radiology and Nuclear Medicine, MHeNs - Mental Health and Sciences Research Institute Maastricht University (Medical Centre), Maastricht, The Netherlands, [48]Department of Neuroradiology, Christiana Care, Pennsylvania, USA, [49]Department of Radiology, Tribhuvan University Teaching Hospital, Kathmandu, Bagmati, Nepal, [50]Department of Neuroradiology, University Hospital Tübingen, Tübingen, Germany, [51]Heidelberg University Hospital, Heidelberg, Germany, [52]Boston University Chobanian & Avedisian School of Medicine, Boston, Massachusetts, USA, [53]Uniformed Services University, Bethesda, MD, USA, [54]Department of Radiology, Columbia University Irving Medical Center,

New York, New York, USA, [55]Turku University Hospital and University of Turku, Turku, Finland, [56]Neuroradiology, Lahey Hospital and Medical Center, Burlington, Massachusetts, USA, [57]Mayo Clinic, Jacksonville, Florida, USA, [58]UTSouthwestern Medical Center, Dallas, Texas, USA, [59]University of Pittsburgh Neuroradiology, Pittsburgh, Pennsylvania, USA, [60]Mid Yorkshire Teaching NHS Trust, Wakefield, West Yorkshire, United Kingdom, [61]Department of Radiology, Hospital Universitario La Paz, Madrid, Spain, [62]Yale New Haven Health Bridgeport Hospital, Bridgeport, Connecticut, USA

## Summary

The RSNA Intracranial Aneurysm (RSNA-ICA) Dataset comprises 2566 CTA, 2166 MRA, and 2470 MRI series with expert annotations for intracranial aneurysm presence and location; the publicly released portion is available at https://mira.rsna.org/dataset/7.

## Key Points

- The RSNA Intracranial Aneurysm (RSNA-ICA) Dataset comprises over 7200 CTA, MRA, and MRI series from 21 institutions across five continents, providing broad geographic and imaging diversity.
- The dataset includes imaging series and expert annotations indicating both the presence and location of intracranial aneurysms, with 178 series also including 3D segmentations of the 13 challenge-defined vascular locations.
- RSNA-ICA was used for the 2025 RSNA Intracranial Aneurysm Detection Challenge and is made freely available to the research community for noncommercial use.

## Key Words:

## Abbreviations:

AI – Artificial Intelligence

ASNR – American Society of Neuroradiology

CTA – Computed Tomography Angiography

DICOM – Digital Imaging and Communications in Medicine

ESNR – European Society of Neuroradiology

MRA – Magnetic Resonance Angiography

MRI – Magnetic Resonance Imaging

NIfTI – Neuroimaging Informatics Technology Initiative

RSNA-ICA – Radiological Society of North America Intracranial Aneurysm

SNIS – Society of Neurointerventional Surgery

## Introduction

Intracranial aneurysm rupture can lead to subarachnoid hemorrhage, resulting in substantial morbidity and death (1). Detection can be challenging, particularly for aneurysms smaller than 5 mm or on non-angiographic brain imaging studies, and lesions that go unrecognized may enlarge over time and eventually rupture. Artificial intelligence (AI) has the potential to assist in the detection of intracranial aneurysms, particularly in the opportunistic screening setting, including on non-angiographic imaging studies. Several algorithms have been developed to detect intracranial aneurysms on computed tomography angiography (CTA) (2–5) and magnetic resonance angiography (MRA) (6–8). However, intracranial aneurysms can also be detected on routine brain magnetic resonance imaging (MRI) examinations, which are performed more frequently than MRA. Because vascular structures are often visible on commonly acquired sequences, including T2-weighted and post-contrast T1-weighted images, routine non-angiographic MRI may also enable opportunistic screening. AI could therefore potentially identify incidental aneurysms on routine MRI examinations obtained for other clinical indications. Previously published open CTA and MRA aneurysm datasets have generally been limited to a few hundred studies, and most do not include routine MRI examinations (9–12).

The Radiological Society of North America (RSNA) has hosted annual AI challenges since 2017, providing de-identified, annotated clinical imaging datasets for algorithm development and evaluation against a hidden test set (13). For the 2025 RSNA AI Challenge, the objective was to develop algorithms to detect and localize intracranial saccular aneurysms on CTA, MRA, and MRI.

In collaboration with the American Society of Neuroradiology (ASNR), the Society of Neurointerventional Surgery (SNIS), and the European Society of Neuroradiology (ESNR), RSNA curated a large, publicly available, expert-annotated dataset of CTA, MRA, and MRI series from patients with and without saccular intracranial aneurysms. The RSNA Intracranial Aneurysm (RSNA-ICA) Dataset was used for the 2025 RSNA AI Challenge.

## Dataset Curation and Annotation

An overview of the dataset curation and annotation process is provided in Figure 1, with a detailed description in Appendix S1. All participating sites obtained ethical approval, and either written informed consent or a waiver of consent was obtained. Participating sites provided initial labels for aneurysm presence and location based on clinical reports, using the 13 challenge-defined vascular locations: infraclinoid internal carotid artery (left/right), supraclinoid internal carotid artery (left/right), middle cerebral artery (left/right), anterior communicating artery, anterior cerebral artery (left/right), posterior communicating artery (left/right), basilar tip, and other posterior circulation. Radiologist annotators recruited by the RSNA, ASNR, SNIS, and ESNR independently annotated aneurysm presence and location and placed a crosshair localizer at the center of each aneurysm. Detailed instructions were provided, including written guidelines (supplemental file “RSNA-ICA Annotation Instructions”) with illustrative images and an instructional video. All annotators completed a training module to familiarize themselves with the annotation tools and process. Performance during training was evaluated against reference annotations defined by the organizing committee, and only those who demonstrated acceptable accuracy were selected to annotate the dataset.

The annotator pool consisted of 65 attending radiologists from 20 countries, with a median post-training experience of 6 years (IQR, 3-14 years). Annotations were collected using a crowdsourcing approach, with a public scoreboard to promote participation. Annotators completed a mean of 86.8 cases (SD, 91.6 cases). Discrepancies between site-provided and independent annotations were adjudicated by annotators with the highest relative performance on the initial dataset through review of the images and both sets of labels.

A subset of the dataset (178 series across 105 patients) includes 3D segmentations of the 13 vascular locations (Figure 2). These segmentations were generated using an iterative refinement approach with MONAI Label (14) and a pretrained model based on manually adjusted segmentations from the TopCoW dataset (15), followed by manual review and approval by neuroradiologists from the organizing committee. The segmented cases were selected from patients with intracranial aneurysms, and when visible on the segmented series, the aneurysm was included in the vascular segmentation of its parent vessel.

A total of 6509 image series from 3848 patients were included in the challenge dataset. These were divided into a training set (4348/6509, 66.8%), a public test set (689/6509, 10.6%), and a private test set (1472/6509, 22.6%). The data split accounted for multiple exams/modalities per patient by assigning all exams/modalities from a single patient to the same split; details of the sampling process are provided in Appendix S1. Following post-challenge quality control, two series were removed from the training set because of severe imaging artifacts, resulting in 4346 training series in the final MIRA release. After completion of the challenge, an additional 695 series from 430 patients across three sites not represented in the original challenge dataset (Sites 19–21) were added as an out-of-distribution (OOD) cohort using the same curation and annotation process as the original challenge dataset. The OOD cohort was not part of the challenge itself and is available only through MIRA.

## Dataset Description and Usage

RSNA-ICA contains 2566 CTA, 2166 MRA, and 2470 MRI series from 4278 adult (≥18 years) patients, with a total of 7202 image series from 21 institutions across 12 countries and five continents. A total of 4099 aneurysms were identified in 2261 patients (average of 1.8 aneurysms/patient) and were distributed across the training set (2252 aneurysms), public test set (397 aneurysms), private test set (1093 aneurysms), and post-challenge dataset (357 aneurysms). The distribution of aneurysm location, patient age, sex, and imaging modality is shown in Figure 3. A detailed summary of patient demographics and aneurysm locations is provided in Table 1 by dataset subset and in Table 2 by institution.

Dataset images consistently include the Circle of Willis, with variable coverage of the neck vessels and the distal intracranial vessels reflecting real-world field-of-view selection. The dataset is provided in Digital Imaging and Communications in Medicine (DICOM) format, along with voxel-level segmentations of the challenge-defined arterial segments in Neuroimaging Informatics Technology Initiative (NIfTI) format for 178 series and two CSV annotation files containing series-level labels and aneurysm localizers. Publicly released data are available through MIRA at https://mira.rsna.org/dataset/7, including 4346 series from the original challenge training set and an additional 695-series post-challenge dataset from three previously unseen sites; the original 4348-series challenge training set (without exclusions for image quality) is also available through Kaggle at https://www.kaggle.com/competitions/rsna-intracranial-aneurysm-detection.

Image files are organized into folders by SeriesInstanceUID, a unique series-level identifier. Individual DICOM files within each folder are named using the SOPInstanceUID and can be ordered according to their position within the series.

In the MIRA release, PatientGroupID and StudyGroupID provide anonymized patient- and study-level linkage across series and should be used for grouping. PatientID and StudyInstanceUID are also provided to match the corresponding

DICOM header values; details of identifier handling for the original challenge and OOD cohorts are provided in Appendix S1.

Series-level annotations and demographic information are provided in two CSV files containing clinical labels, localization data, and metadata. File naming differs between the Kaggle and MIRA distributions: on Kaggle, these files are named train.csv and train_localizers.csv; in the MIRA release, the corresponding files are named dataset.csv and dataset_localizers.csv. The primary annotation file contains one row corresponding to each imaging series identified by SeriesInstanceUID and includes aneurysm presence and location labels, along with associated demographic information. A value of 0 indicates absence and 1 indicates presence of an aneurysm at a given arterial location.

The localizer file provides the localization of individual aneurysms. The SOPInstanceUID identifies the corresponding DICOM image and can be combined with ".dcm" to generate the filename. The coordinates specify the location of the aneurysm within the image.

Voxel-level segmentations of the arterial locations are provided in NIfTI format for a subset of 178 series from the training set. Files are named using SeriesInstanceUID, and each segmentation contains separate labels for the 13 arterial locations. Examples of aneurysms visualized on CTA, MRA, and MRI, along with their corresponding arterial segmentations, are presented in Figure 2.

## Discussion

We curated a large, high-quality intracranial CTA, MRA, and MRI dataset with expert annotations from 21 institutions across five continents, representing the largest and most diverse public dataset of intracranial aneurysms currently available.

Participating sites were asked to enrich the dataset with rarer aneurysm locations, although some locations remained less common than others. To address this imbalance and ensure sufficient training examples, we combined posterior circulation arterial locations other than the basilar tip into a single label. We provided the data in DICOM format and included images with heterogeneous acquisition and reconstruction parameters to better reflect variability encountered in routine clinical practice. Care was also taken to balance the training and test sets with respect to imaging modality, contributing site, and aneurysm location. This design was intended to support model robustness across heterogeneous imaging data and evaluation settings. A limitation is the absence of annotations for aneurysm rupture or subarachnoid hemorrhage, which likely had a higher prevalence in the aneurysm-positive group; this imbalance may introduce bias, as the presence of hemorrhage on imaging could influence the models toward predicting aneurysms.

As in previous RSNA challenges (16–18), we aimed to maximize data utility and ensure high-quality labels while minimizing annotator burden. Contributing sites provided initial labels based on clinical reports. Annotators independently reviewed and labeled the same cases without access to the site-provided labels. In cases of disagreement, a third adjudication review was performed with both annotations visible to resolve the discrepancy.

While multiple measures were implemented to ensure annotation consistency and accuracy, some limitations remain. Small aneurysms (<3 mm) were particularly challenging to label, especially when differentiation from an arterial infundibulum was uncertain, reflecting a real-world clinical challenge. Cases with uncertain or discordant initial annotations underwent adjudication with review of the imaging and available labels; an aneurysm was retained as positive only when the adjudicator was sufficiently confident that a true aneurysm was present. Cases that remained indeterminate after adjudication were classified as negative to reduce false positive annotations. This conservative approach may have resulted in some false-negative labels for very small or equivocal lesions. Precise localization was limited by the study design, in which individual vessel segments were grouped together (*e.g.,* clinoid, ophthalmic, and supraclinoid internal carotid artery segments were all designated supraclinoid), and multiple posterior circulation locations were combined into a single label (*e.g.,* vertebral, basilar trunk, posterior cerebral, and cerebellar artery locations). The grouping of internal carotid artery segments was intended to distinguish predominantly intradural from extradural locations, which differ in their potential to cause subarachnoid hemorrhage, although this distinction is not always possible on noninvasive imaging. Precise localization was also difficult when aneurysms arose at the border between vessel segments (*e.g.,* carotid terminus aneurysms). Moreover, annotations were performed on a web-based platform, which does not fully replicate clinical practice, in which radiologists have access to high-resolution monitors, multiplanar thin-section imaging, clinical context, and prior examinations.

In summary, RSNA-ICA represents the largest and most diverse publicly available expert-annotated collection of CTA, MRA, and MRI series for intracranial aneurysm detection. With the release of this dataset, we aim to advance machine learning research in neurovascular imaging and support future improvements in diagnostic accuracy, efficiency, and patient outcomes. This dataset is freely available to researchers for noncommercial use.

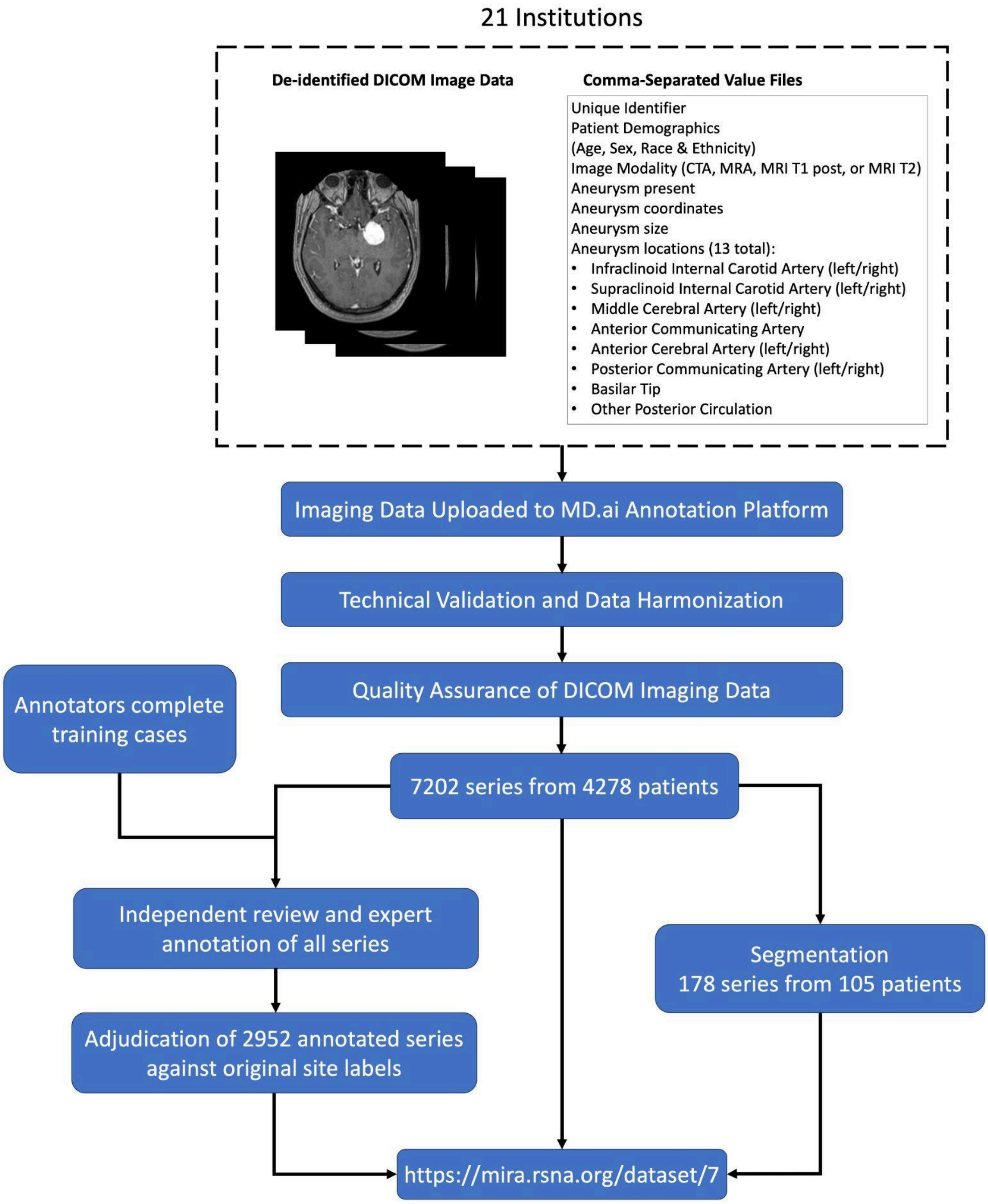

Figure 1. Flowchart illustrating the data curation and annotation workflow. DICOM = Digital Imaging and Communications in Medicine

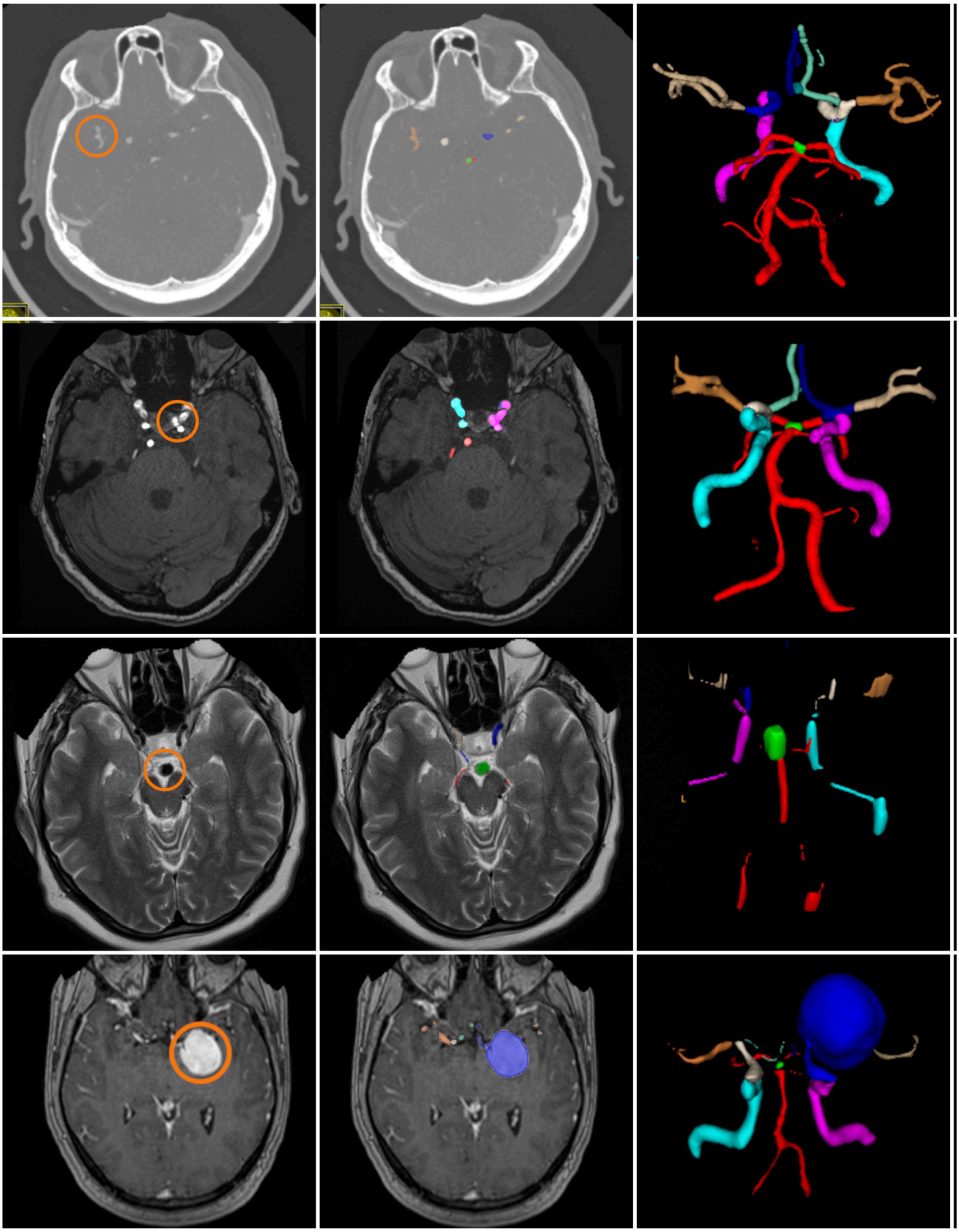

Figure 2. Examples of vessel segmentation, with each color corresponding to a specific vascular location. The left column shows the original images with aneurysms encircled in orange, the middle column shows the vessel segmentation overlays, and the right column shows 3D reconstructions. The top row shows a right middle cerebral artery aneurysm on CTA; the second row, a left infraclinoid internal carotid artery aneurysm on MRA; the third row, a basilar tip aneurysm on T2-weighted MRI; and the bottom row, a left supraclinoid internal carotid artery aneurysm on post-contrast T1-weighted MRI.

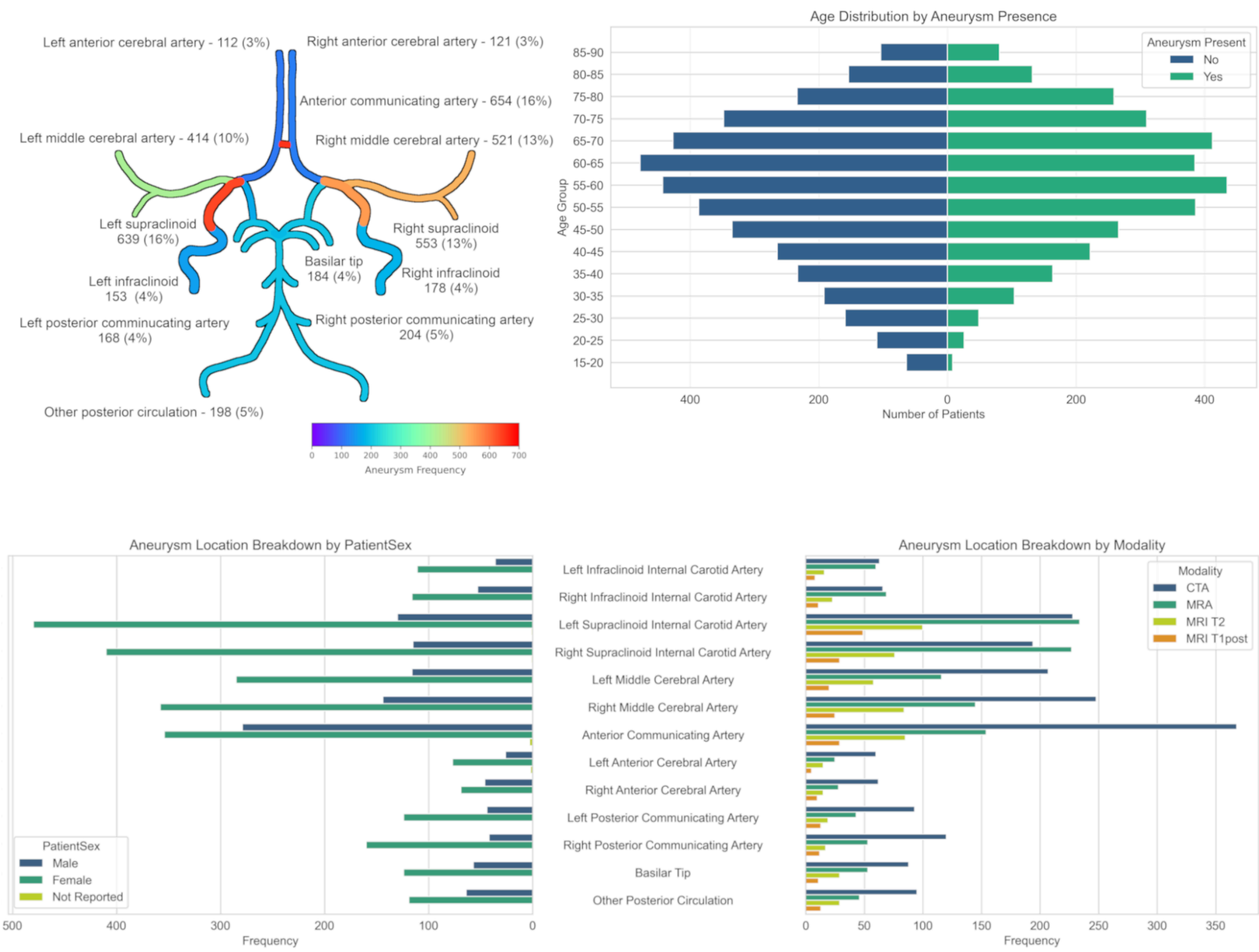

Figure 3. Distribution of intracranial aneurysms by location, age, sex, and imaging modality. Top left: Schematic illustration depicting the frequency of aneurysms across different anatomical locations. Top right: Age distribution of the study population displayed as a mirrored horizontal bar chart. Bars on the left represent patients without aneurysms, and bars on the right represent patients with aneurysms. Patient age was not reported in 30 series. Bottom left: Horizontal bar chart illustrating aneurysm location according to patient sex. Patient sex was not reported in 18 series. Bottom right: Horizontal bar chart illustrating aneurysm location according to imaging modality.

Table 1. Demographic distribution and number of positive and negative series for aneurysms by dataset subset

| Data split | Sex* | | Age** (y), mean ± SD (range) | CTA | MRA | MRI T1 post | MRI T2 | Total Series | Negative Series | Positive Series | Total Aneurysms | Aneurysm Size ≥3 mm | Aneurysm Location Distribution*** | | | | | | | | | | | | |
|---|---|---|---|---|---|---|---|---|---|---|---|---|---|---|---|---|---|---|---|---|---|---|---|---|---|
| | Male | Female | | | | | | | | | | | 1 | 2 | 3 | 4 | 5 | 6 | 7 | 8 | 9 | 10 | 11 | 12 | 13 |
| Training | 908 | 1929 | 58.5 +/-15.7 (18-89) | 1808 | 1251 | 305 | 982 | 4346 | 2482 | 1864 | 2252 | 1627 | 82 | 105 | 343 | 289 | 222 | 307 | 376 | 44 | 63 | 85 | 103 | 111 | 122 |
| Public test | 180 | 208 | 57.5 +/-15.4 (18-89) | 151 | 257 | 60 | 221 | 689 | 357 | 332 | 397 | 253 | 16 | 18 | 60 | 53 | 33 | 42 | 69 | 11 | 15 | 20 | 25 | 13 | 22 |
| Private test | 303 | 320 | 59.2 +/-15.7 (18-89) | 342 | 426 | 284 | 420 | 1472 | 699 | 773 | 1093 | 640 | 41 | 47 | 183 | 168 | 113 | 130 | 143 | 36 | 35 | 46 | 61 | 48 | 42 |
| Post-challenge data set | 172 | 243 | 54.0 +/-15.1 (18-89) | 265 | 232 | 58 | 140 | 695 | 406 | 289 | 357 | 320 | 14 | 8 | 53 | 43 | 46 | 42 | 66 | 21 | 8 | 17 | 15 | 12 | 12 |
| Total | 1563 | 2700 | 58.1 +/-15.7 (18-89) | 2566 | 2166 | 707 | 1763 | 7202 | 3944 | 3258 | 4099 | 2840 | 153 | 178 | 639 | 553 | 414 | 521 | 654 | 112 | 121 | 168 | 204 | 184 | 198 |

* Sex not reported in 15 patients in the “Post-challenge data set”

** Age not reported in 18 patients in the “Post-challenge data set”

*** Aneurysm location was based on the 13 challenge-defined vascular locations: 1 = left infraclinoid internal carotid artery; 2 = right infraclinoid internal carotid artery; 3 = left supraclinoid internal carotid artery; 4 = right supraclinoid internal carotid artery; 5 = left middle cerebral artery; 6 = right middle cerebral artery; 7 = anterior communicating artery; 8 = left anterior cerebral artery; 9 = right anterior cerebral artery; 10 = left posterior communicating artery; 11 = right posterior communicating artery; 12 = basilar tip; and 13 = other posterior circulation.

Table 2. Demographic distribution and number of positive and negative series for aneurysms by institution

| Site ID | Sex* | | Age** (y), mean ± SD (range) | CTA | MRA | MRI T1 post | MRI T2 | Total Series | Negative Series | Positive Series | Total Aneurysms | Aneurysm Size ≥3 mm | Aneurysm Location Distribution*** | | | | | | | | | | | | |
|---|---|---|---|---|---|---|---|---|---|---|---|---|---|---|---|---|---|---|---|---|---|---|---|---|---|
| | Male | Female | | | | | | | | | | | 1 | 2 | 3 | 4 | 5 | 6 | 7 | 8 | 9 | 10 | 11 | 12 | 13 |
| Site 1 | 88 | 82 | 51.0 +/- 15.0 (19-89) | 189 | 36 | 13 | 23 | 261 | 101 | 160 | 195 | 166 | 1 | 5 | 21 | 28 | 12 | 16 | 59 | 3 | 17 | 5 | 11 | 6 | 11 |
| Site 2 | 74 | 125 | 69.8 +/- 14.9 (21-89) | 99 | 100 | 96 | 98 | 393 | 204 | 189 | 246 | 175 | 14 | 13 | 43 | 20 | 17 | 21 | 22 | 4 | 4 | 26 | 25 | 9 | 28 |
| Site 3 | 103 | 94 | 61.4 +/- 15.2 (18-89) | 110 | 122 | 0 | 106 | 338 | 222 | 116 | 130 | 72 | 9 | 2 | 33 | 26 | 6 | 4 | 16 | 3 | 3 | 6 | 6 | 4 | 12 |
| Site 4 | 69 | 196 | 57.1 +/- 17.0 (18-89) | 106 | 150 | 130 | 151 | 537 | 359 | 178 | 213 | 136 | 12 | 16 | 54 | 41 | 10 | 15 | 29 | 1 | 3 | 5 | 9 | 9 | 9 |
| Site 5 | 55 | 144 | 47.4 +/- 17.1 (19-89) | 103 | 65 | 37 | 159 | 364 | 251 | 113 | 135 | 86 | 7 | 7 | 31 | 24 | 10 | 25 | 8 | 3 | 3 | 1 | 4 | 10 | 2 |
| Site 6 | 107 | 167 | 65.0 +/- 14.6 (18-89) | 140 | 160 | 7 | 11 | 318 | 131 | 187 | 230 | 127 | 4 | 7 | 22 | 32 | 30 | 45 | 41 | 4 | 6 | 11 | 9 | 13 | 6 |
| Site 7 | 70 | 129 | 55.2 +/- 13.7 (19-87) | 184 | 77 | 7 | 21 | 289 | 13 | 276 | 371 | 280 | 14 | 4 | 34 | 31 | 70 | 74 | 71 | 15 | 12 | 12 | 15 | 7 | 12 |
| Site 8 | 82 | 138 | 59.0 +/- 16.0 (21-89) | 22 | 197 | 34 | 127 | 380 | 236 | 144 | 175 | 99 | 8 | 12 | 25 | 29 | 25 | 22 | 25 | 2 | 4 | 4 | 6 | 5 | 8 |
| Site 9 | 52 | 77 | 64.5 +/- 15.6 (25-89) | 131 | 0 | 0 | 0 | 131 | 63 | 68 | 88 | 61 | 3 | 7 | 12 | 3 | 14 | 8 | 19 | 2 | 1 | 3 | 7 | 6 | 3 |
| Site 10 | 57 | 155 | 56.4 +/- 13.5 (19-81) | 161 | 91 | 19 | 85 | 356 | 163 | 193 | 261 | 173 | 10 | 6 | 36 | 33 | 38 | 50 | 33 | 7 | 5 | 3 | 13 | 12 | 15 |
| Site 11 | 68 | 179 | 55.7 +/- 13.9 (18-89) | 0 | 241 | 0 | 253 | 494 | 246 | 248 | 261 | 160 | 4 | 8 | 61 | 50 | 30 | 46 | 37 | 4 | 5 | 6 | 5 | 0 | 5 |
| Site 12 | 85 | 113 | 54.2 +/- 14.7 (19-84) | 129 | 68 | 24 | 62 | 283 | 160 | 123 | 170 | 127 | 4 | 7 | 14 | 11 | 26 | 16 | 36 | 8 | 1 | 13 | 17 | 6 | 11 |
| Site 13 | 67 | 119 | 58.9 +/- 15.7 (18-89) | 92 | 99 | 0 | 1 | 192 | 93 | 99 | 120 | 74 | 4 | 11 | 21 | 13 | 14 | 20 | 21 | 7 | 1 | 0 | 3 | 2 | 3 |
| Site 14 | 82 | 125 | 61.4 +/- 16.3 (20-89) | 204 | 107 | 77 | 42 | 430 | 263 | 167 | 252 | 167 | 13 | 14 | 27 | 30 | 11 | 41 | 39 | 6 | 9 | 17 | 22 | 13 | 10 |
| Site 15 | 55 | 130 | 61.9 +/- 15.7 (20-89) | 104 | 88 | 65 | 45 | 302 | 174 | 128 | 168 | 127 | 2 | 22 | 50 | 25 | 11 | 17 | 15 | 5 | 3 | 5 | 4 | 5 | 4 |
| Site 16 | 99 | 169 | 67.6 +/- 13.2 (18-89) | 184 | 95 | 93 | 269 | 641 | 431 | 210 | 240 | 137 | 15 | 14 | 43 | 32 | 9 | 31 | 28 | 6 | 3 | 12 | 13 | 7 | 27 |
| Site 17 | 59 | 136 | 54.1 +/- 13.1 (19-89) | 127 | 65 | 2 | 1 | 195 | 94 | 101 | 127 | 90 | 3 | 7 | 24 | 30 | 8 | 5 | 16 | 5 | 2 | 9 | 4 | 9 | 5 |
| Site 18 | 119 | 179 | 58.7 +/- 13.7 (22-89) | 216 | 173 | 45 | 169 | 603 | 334 | 269 | 360 | 263 | 12 | 8 | 35 | 52 | 27 | 23 | 73 | 6 | 31 | 13 | 16 | 49 | 15 |
| Site 19 | 81 | 104 | 56.6 +/- 14.6 (18-86) | 133 | 102 | 47 | 48 | 330 | 177 | 153 | 198 | 182 | 13 | 5 | 33 | 23 | 27 | 20 | 33 | 7 | 3 | 10 | 8 | 9 | 7 |
| Site 20 | 52 | 103 | 52.8 +/- 15.8 (20-89) | 68 | 91 | 4 | 76 | 239 | 167 | 72 | 88 | 75 | 1 | 0 | 2 | 8 | 13 | 20 | 19 | 4 | 3 | 4 | 7 | 3 | 4 |
| Site 21 | 39 | 36 | 50.2 +/- 13.5 (19-86) | 64 | 39 | 7 | 16 | 126 | 62 | 64 | 71 | 63 | 0 | 3 | 18 | 12 | 6 | 2 | 14 | 10 | 2 | 3 | 0 | 0 | 1 |

* Sex not reported in total 15 patients from site 19 and 21

** Age not reported in total 18 patients from site 19 and 21

*** Aneurysm location was based on the 13 challenge-defined vascular locations: 1 = left infraclinoid internal carotid artery; 2 = right infraclinoid internal carotid artery; 3 = left supraclinoid internal carotid artery; 4 = right supraclinoid internal carotid artery; 5 = left middle cerebral artery; 6 = right middle cerebral artery; 7 = anterior communicating artery; 8 = left anterior cerebral artery; 9 = right anterior cerebral artery; 10 = left posterior communicating artery; 11 = right posterior communicating artery; 12 = basilar tip; and 13 = other posterior circulation.

# Appendix

## Data Collection

RSNA-ICA comprises CTA, MRA, MRI T2, and MRI T1 post images obtained at 21 institutions across five continents. Each contributing institution was asked to retrospectively identify cases, with at least half being positive for one or more intracranial saccular aneurysms (cases with fusiform, mycotic, or pseudoaneurysms, or arteriovenous malformations were all classified as negative for aneurysms unless a flow-related saccular aneurysm was present outside the arteriovenous malformation nidus). The remaining cases included similar imaging studies without aneurysms, although other non-aneurysmal pathologies were allowed. Previously treated aneurysms were included in negative cases if no residual or recurrent aneurysm was present. Imaging data included CTA and/or MRA (time-of-flight MRA [TOF-MRA] or post-contrast MRA), with efforts to balance both modalities. Brain MRI sequences (MRI T2 and MRI T1 post) were included for at least 50% of patients. For positive cases, these sequences were included only if the aneurysm was visible. Each modality was limited to a single time point per patient, with no strict time frame required between CTA/MRA and MRI, provided no interval treatment had occurred. The dataset was curated rather than intended to reflect real-world prevalence, with sites asked to provide approximately 50% positive cases and include aneurysms in less common locations. This balanced design was intended to reduce extreme class imbalance, facilitate comparison across teams, and expose models to a broader spectrum of aneurysm presentations than would be seen in routine clinical practice. Cases with severe motion artifacts or other technical artifacts limiting diagnostic evaluation were excluded.

A total of 6509 series from 3848 patients were included in the challenge dataset. These were divided into a training set (4348/6509, 66.8%), a public test set (689/6509, 10.6%), and a private test set (1472/6509, 22.6%). The segmentation subset was fully included in the training set, and approximately 30% of the remaining series were randomly selected for the test sets. Stratified sampling by modality, site, and aneurysm location ensured balanced representation in the test sets. Additional positive cases were oversampled from male patients, those aged 18–35 years, and cases with multiple or small aneurysms (<3 mm), as well as from basilar tip aneurysms and underrepresented contributing sites, to ensure adequate diversity and inclusion of challenging examples. The private test set was sampled from the full test set using the same procedure. When multiple modalities were available from a single patient, all modalities were assigned to a single split. The 4348 series in the original challenge training set were publicly released on Kaggle, whereas the 2161 series in the public and private test sets remain withheld to facilitate unbiased evaluation of new automated detection methods. During post-challenge quality control, two training series were removed because of severe imaging artifacts, leaving 4346 series from the original challenge training set in the final public release. The affected patients and studies remain represented by other series in the dataset. Following completion of the challenge, an additional 695 series across three sites not represented in the original challenge dataset were incorporated into the dataset as a post-challenge dataset using the same curation and annotation process as the original challenge dataset. Thus, the MIRA release includes 5041 publicly available series (4346 retained challenge-training series and 695 post-challenge series).

## Data preprocessing

Imaging data were de-identified using the RSNA CTP DICOM Anonymizer (https://mircwiki.rsna.org/index.php?title=The_CTP_DICOM_Anonymizer). All images underwent defacing using a deep learning–based method designed to remove identifying facial and auricular features while preserving intracranial and vascular anatomy. The defacing method used a three-dimensional U-Net implemented within the nnU-Net framework to segment facial and auricular regions for removal. The model was developed using fivefold cross-validation on a cohort of 425 imaging series from 425 unique individuals, comprising 255 individuals from Cleveland Medical Center and 170 from Fleni, Buenos Aires. For DICOM input, the software generated a defaced DICOM series while retaining the series structure, facilitating integration into clinical and research imaging workflows. Details for the defacing method development and evaluation will be reported in a separate publication.

## Technical and Visual Validation

For the imaging dataset, DICOM unique identifiers, pixel geometry parameters, and private tags were validated against the DICOM standard using a metadata processing tool. Custom Python scripts were developed using the pydicom library (1) to extract and manage metadata, following approaches similar to previous RSNA challenges. A comprehensive list of unique tags and values was reviewed, with particular attention to Patient ID, StudyInstanceUID, SeriesInstanceUID, SOPInstanceUID, SOPClassUID, and modality-specific acquisition tags for CT and MR. Automated checks were performed at multiple stages to assess DICOM compliance, data consistency across sources, and the absence of protected health information. Initial curation and PHI removal were performed using the RSNA CTP DICOM Anonymizer, which follows HIPAA “safe harbor” guidelines. A secondary verification step using custom Python scripts provided additional validation of the DICOM metadata. Files that failed these checks were quarantined for review and removed when necessary.

For the challenge release (Sites 1–18), patient- and study-level identifiers within the DICOM headers were blinded at the series level, thereby preventing linkage of multiple series from the same patient or study. These DICOM header values are retained unchanged in the MIRA release and are also provided in the CSV files as PatientID and StudyInstanceUID. The OOD cohort (Sites 19–21) was also de-identified but preserves the patient–study–series structure supplied by the contributing sites. To enable linkage across the full release, the MIRA CSV files additionally include anonymized PatientGroupID and StudyGroupID fields identifying series from the same patient and study, respectively, built for each site from that site's own records and validated against the DICOM metadata. Patient- and study-level grouping of the MIRA data should therefore be performed using PatientGroupID and StudyGroupID rather than the DICOM header identifiers.

Only essential DICOM information was retained in the curated dataset, including pixel data, image geometry, acquisition parameters, device information, necessary image- and series-level identifiers, and basic patient information such as age and sex. All other private or non-essential tags were removed. In addition to technical validation, radiologists performed visual review to ensure image quality. Studies that were incomplete or contained series outside the project scope were flagged and removed using md.ai annotation labels.

## Annotation

Sites labeled aneurysm presence from clinical reports and assigned locations based on challenge-defined vascular locations:

1. Left infraclinoid internal carotid artery (including lacerum, petrous, or cavernous segments)
2. Right infraclinoid internal carotid artery (including lacerum, petrous, or cavernous segments)
3. Left supraclinoid internal carotid artery (including clinoid, ophthalmic, or supraclinoid segments)
4. Right supraclinoid internal carotid artery (including clinoid, ophthalmic, or supraclinoid segments)
5. Left middle cerebral artery (any segment)
6. Right middle cerebral artery (any segment)
7. Anterior communicating artery
8. Left anterior cerebral artery (any segment)

9. Right anterior cerebral artery (any segment)
10. Left posterior communicating artery
11. Right posterior communicating artery
12. Basilar tip
13. Other posterior circulation: posterior circulation excluding the basilar tip (e.g., vertebral artery, remaining basilar artery, posterior inferior cerebellar artery, anterior inferior cerebellar artery, superior cerebellar artery or posterior cerebral artery)

The classification of aneurysm location was based on the vessel of origin. For aneurysms arising at a side branch, the aneurysm was classified as originating from the side branch artery (e.g., an aneurysm at the junction of the internal carotid artery and posterior communicating artery was labeled as posterior communicating artery). Radiologist annotators recruited through RSNA, in collaboration with ASNR, SNIS, and ESNR, independently annotated aneurysm presence and location, including placement of a crosshair marker at the aneurysm, using a web-based annotation platform (md.ai, New York). All annotators were required to complete a training module and demonstrate acceptable accuracy on 15 test cases before contributing to the dataset.

Annotators used the best diagnostic CTA or MRA to determine aneurysm presence and location based on the challenge-defined vascular locations. For positive cases, the label was applied to MRI only if the aneurysm was clearly visible; otherwise, the MRI series was included among the negative cases. During annotation, candidate aneurysms were categorized as possible small aneurysm (<3 mm) versus infundibulum, definite small aneurysm (<3 mm), or definite aneurysm (≥3 mm). When an independent annotation of a possible small aneurysm versus infundibulum matched the site-provided aneurysm label, the lesion was retained as an aneurysm. Ambiguous cases and disagreements between site-provided and independent annotations were reviewed by expert neurointerventional adjudicators or neuroradiologists from the organizing committee using the md.ai platform; in total, 2952 series were adjudicated. Adjudicators were instructed to retain an aneurysm only when sufficiently confident that a true aneurysm was present. Lesions that remained indeterminate after adjudication were classified as negative and are not represented by aneurysm localizers in the released dataset. For retained aneurysms, the MIRA dataset provides the size category (<3 mm or ≥3 mm).

Annotators also indicated the image type: CTA, MRA, MRI T2 (including FIESTA/CISS), or post-contrast T1-weighted MRI. The full annotation protocol for aneurysm presence and location is available in the supplemental file “RSNA-ICA Annotation Instructions.” In summary, the dataset includes binary labels indicating the presence or absence of aneurysm at each anatomical location for each imaging series, as well as a localizer marking the aneurysm within the image.

A subset of the training set (178 series across 105 patients) includes 3D segmentations of the vascular locations. These segmentations were generated using MONAI Label (2) and a pretrained model based on manually adjusted segmentations from the TopCoW dataset (3). The model was iteratively retrained as new cases were added. The segmented cases were selected from patients with intracranial aneurysms, and when visible on the segmented series, the aneurysm was included in the vascular segmentation of its parent vessel. All segmentations underwent manual review and approval by neuroradiologists from the organizing committee. The full annotation protocol for the vessel segmentations is available in the supplemental file “RSNA-ICA Segmentation Instructions”.

## Study Identification, Data Extraction, and Data De-Identification

Contributing sites are listed in alphabetical order and do not correspond to the order in Table 2.

### *Aga Khan University Hospital, Karachi, Pakistan*

The Aga Khan University site investigator conducted a search using in built PACS search tool for MRA and CTA of the brain performed between January 1, 2010, and December 1, 2024. The search was restricted to patients aged 18 years or older, using key words “intracranial aneurysm”. This search retrieved approximately 767 cases. Out of these, a total of 98 positive aneurysm cases were randomly selected from the list. For negative cases, the keywords were: “No intracranial aneurysm or AVM” and the PACS search was performed between January 1, 2023- Feb 15, 2025. This retrieved a total of 500 cases, out of which 86 negative cases were included, randomly selected from the beginning of the list. Each examination was manually reviewed to confirm that inclusion criteria were met, and cases were excluded according to the exclusion criteria. A total of 184 examinations with 290 series were finally selected. Most of our positive cases were 3 mm axial, coronal and sagittal sections, predominantly 3mm axials for CTAs, 5mm for T2 axial or Sagittal MRI, and source data for TOF MRA and T1 post contrast MRI. However, most of the negative cases were source data for CTAs and MRAs. At least three of our negative cases were positive for AVMs, (Mentioned in the data sheet).

For each positive case, the aneurysm location was recorded. For each negative case, a second review confirmed that no aneurysms were visualized. Patient age and sex at birth were also documented. All selected MR studies were downloaded from PACS to a local workstation within the hospital network using RSNA Anonymizer (Radiological Society of North America). All the anonymized series were manually identified within each study and uploaded to MD.ai for annotation.

### *Big Data Center, China Medical University Hospital, Taiwan*

China Medical University Hospital (CMUH) team identified patients who underwent brain angiography, including computed tomography (CTA) and magnetic resonance angiography (MRA), between 2008 and 2022 using the iHi Data Platform. This platform contains carefully verified electronic medical records from more than 3.3 million patients who have received healthcare services at CMUH (4).

To identify aneurysm cases, we initially included adult patients (aged ≥18 years) with a diagnosis of cerebral aneurysm (defined by ICD-9-CM code 437.3 or ICD-10-CM code I67.1), with aneurysm-related keywords documented in their CTA or MRA reports, and without a history of transcatheter arterial embolization (TAE) or craniotomy. Following manual image annotation and expert review, 100 patients with confirmed cerebral aneurysmal lesions were selected from 358 eligible patients as the aneurysm case group.

For the control group, we included patients who underwent brain angiography but had no evidence of cerebral aneurysm. Exclusion criteria included any diagnosis related to cerebral aneurysm or intracranial hemorrhage (ICD-9-CM codes 437.3, 430–432, 442.89, or 442.9; ICD-10-CM codes I67.1, I60–I62, or I72.9), the presence of aneurysm-related keywords in radiology reports, or a history of craniotomy or TAE. From 1,072 eligible individuals without aneurysm, 100 were randomly selected as controls.

CTA DICOM series with slice thickness ≤ 5.0 mm and MRA DICOM series with slice thickness ≤ 1.6-mm were retrieved from CMUH's PACS (INFINITT PACS). All images underwent de-identification through the ISO-certified data security pipeline of CMUH, compliant with ISO 29100, ISO 29191, and ISO 27001 standards. Personally identifiable information contained within the DICOM files was irreversibly transformed using hash-based indexing. The final DICOM files retained 42 imaging-related parameters associated with acquisition settings, such as kilovoltage peak (kVp) and milliampere-seconds (mAs), to preserve radiologic dose and imaging context information. CTA and MRA source series (i.e., T2, TOF MRA) were manually identified for each study and uploaded to MD.ai.

This study was approved by the Research Ethics Committee of CMUH (IRB No. CMUH113-REC1-163). The requirement for informed consent was waived due to the retrospective nature of the study.

### *Clinical Center University of Sarajevo, Sarajevo, Bosnia and Herzegovina*

A comprehensive retrospective search was conducted using institutional Radiology Information System and PACS (SECTRA) by institutional procedural codes for brain CTA and MRA. The search included scans acquired between January 2022 and September 2024.

Each examination was manually reviewed to confirm that the inclusion criteria were met. Rigorous peer-reviewed analysis of CTA and MRA reports was undertaken to apply exclusion criteria, which included patients less than 18 years of age and severe motion artifacts. Presence of fusiform or mycotic aneurysm, pseudoaneurysm or AVM (unless flow related aneurysm outside the nidus in AVM) was considered exclusion criteria for positive cases. This meticulous process involved assessment by two experienced radiologists as well as the site's primary investigator. Demographic data were recorded for all cases, and for each positive case identified additional data regarding aneurysm location, rupture status and treatment history. Considering whether CTA or MRA was performed, adequate study series were exported from the PACS system to a local workstation. For patients who had both CTA and MRA, both studies were included. All studies were anonymized using the DICOM anonymizer tool provided by RSNA, and then uploaded to the platform.

### *Duke University, Durham, NC, USA*

Head imaging studies performed for evaluation of suspected or known intracranial aneurysm were retrospectively identified at Duke, together with matched controls. The cohort emphasized cerebrovascular imaging used for aneurysm assessment, including CT angiography (CTA), MR angiography (MRA), and supporting MR sequences such as T2 and T1 post-contrast. Case-level labeling followed the RSNA challenge schema, with Positive indicating aneurysm present and Negative indicating no aneurysm; Negative cases were further sub-typed as Control – Other Pathologies or Treated Aneurysm with Complete Occlusion, consistent with challenge definitions. Radiology free-text report content was used to support label derivation via an in-house validated locally deployed open-weight LLM-assisted extraction system (5), followed by manual review for verification and consistency with the challenge criteria adherence.

Data were organized at the patient, study, and series levels, and DICOM metadata were used to characterize series (Modality, SeriesDescription, ImageType, SliceThickness, and number of images when available). Prior to filtering, the cohort comprised 416 patients, 420 studies, and 2,824 series (MR 1,527; CT 1,297). The curation objective was to retain diagnostically relevant source acquisitions while excluding non-diagnostic or derived series. Series whose descriptions indicated T2, T1 post-contrast, MRA, or CTA were included using case-insensitive text matching; absence of a series description did not, by itself, result in exclusion. Series with ImageType indicating SECONDARY, DERIVED, or REFORMATTED were excluded, as were series with fewer than 10 images; for CT, series with slice thickness ≥2.0 mm were excluded. When required metadata were missing or unreadable, conservative handling was applied and series were not excluded solely on that basis.

Following curation and labeling, the Duke contribution comprised 278 patients and 695 curated series: 152 patients (264 series) were categorized as Positive and 126 patients (431 series) as Negative. Within the Negative group, 69 patients (325 series) were categorized as Control – Other Pathologies and 57 patients (106 series) as Treated Aneurysm with Complete Occlusion. LLM-assisted extraction was performed locally using an on-device backend system using the Phi-4 open-weight model, to derive structured report-based labels and descriptors, including aneurysm presence, aneurysm count, largest reported diameter, anatomical location, treatment status/type and outcome, aneurysm type, modality mentions, and artifact mentions. Outputs were constrained to a predefined schema, and raw and cleaned outputs were retained internally to support auditability.

### *Faculty of Medicine, Chiang Mai University, Chiang Mai, Thailand*

With approval from our institutional review board, a retrospective search was performed using RIS (Envision.Net) to identify CTA and MRA of the brain and neck obtained between January 1st, 2017, and October 31st, 2024. Searching keywords included “aneurysm”, “intracranial aneurysm”, “saccular aneurysm”, “ruptured aneurysm” and “unruptured aneurysm”. Each study was manually reviewed. Studies meeting exclusion criteria provided by the RSNA were excluded. Patient ID of each report was anonymized. Images were downloaded from PACS (Synapse Radiology PACS version 5.7.000; FUJIFILM Medical systems) in DICOM format and de-identified using RSNA Anonymizer. Demographic data including age, sex and self-reported race and ethnicity (if available) was also collected.

### *Fleni, Buenos Aires, Argentina*

A retrospective search was conducted using the institutional electronic medical record system SIDCA (Sistema Integrado de Documentación Clínica Avanzada) to identify brain CTA or MRI examinations relevant to the RSNA Intracranial Aneurysm Detection Challenge, acquired between November 1, 2014, and November 1, 2024. Radiology reports were queried using aneurysm-related keywords and manually reviewed by a neuroradiologist to confirm eligibility. Duplicate examinations and studies with incomplete imaging data were excluded. Demographic information, imaging modality, and aneurysm localization (when present) were recorded in the RSNA Data Submission Form. Eligible studies were retrieved from the institutional PACS and exported in DICOM format using RadiAnt DICOM Viewer (version 2025.1). Only the series required by the challenge specifications were included, and all datasets were visually checked for correct series selection and adequate image quality. All examinations were de-identified locally at Fleni using the RSNA Anonymizer (version 17.1.1). A final manual review confirmed the absence of residual private information and verified DICOM integrity before the datasets were transferred to the RSNA AI Challenge organizing team.

### Gold Coast, Australia

A retrospective report search on Karisma (Radiology Information System Version 4.0.0.569/5975) was performed for MRI brain angiograms and CT brain angiograms performed between January 2019 and November 2024 that included the words "saccular" and "aneurysm". A second retrospective report search on Karisma was performed for MRI brain angiograms (MRA) and CT brain angiograms (CTA) performed from January 2023 to November 2024. Scans were manually reviewed and sorted according to the RSNA “positive” and “negative” case criteria for saccular aneurysms. All studies were reviewed to exclude patients under 18 years of age, or had severe motion artifacts that limited diagnostic evaluation for the radiologist. All relevant data including patient age at time of scan, sex, self reported race and ethnicity (if available), location of aneurysm (if present) were recorded on the RSNA submission spreadsheet template. Time of flight (TOF) MRAs axial (0.5mm) and CTAs axial soft thins (1.0mm) were downloaded from intelePACS (Version 5-6-1-P238) and anonymised via the DICOM anonymizer tool. Anonymised data and submission spreadsheet was submitted to RSNA.

### Hacettepe University School of Medicine, Ankara, Turkey

A comprehensive retrospective search was conducted through the Hacettepe University institutional Radiology Information System (RIS) to identify head and neck Computed Tomography Angiography (CTA) examinations performed between January 1, 2021, and December 31, 2022. Each examination was manually audited by a board-certified radiologist; diagnostic reports and primary image datasets were reviewed to categorize findings as positive or negative for intracranial aneurysm. For the identified cohort, concurrent Brain Magnetic Resonance Imaging (MRI) studies—specifically including Time-of-Flight (TOF) MRA and axial T2-weighted imaging (WI) acquired within the same period—were cross-referenced and retrieved. All imaging data were subjected to a rigorous de-identification process using the Syngo.via (Siemens Medical Solutions, USA) platform to ensure patient privacy. The resulting anonymized multimodality dataset (CTA and MRI/MRA) was then integrated into the study database for subsequent analysis.

### Hospital Regional Universitario de Málaga Malaga, Spain

At Hospital Regional Universitario de Málaga, a retrospective search was performed using the institutional radiology reporting system (Pruebas Diagnósticas por Imagen; PDI DIRAYA). This search included CT angiography (CTA) of the supraaortic arteries or circle of Willis, MR angiography (MRA) of the supraaortic arteries or circle of Willis and MRI of the head with and without contrast media, obtained between January 2015 to December 2024. Keywords for the radiology report search were “aneurysm”, "circle of Willis", "TOF" (Time Of Flight) and “subarachnoid hemorrhage”. All reports were manually reviewed and classified as positive or negative for aneurysms, and all images were reviewed to exclude unclear cases.

Thin-section (maximum, 1.0 mm) axial soft tissue window reconstructions were downloaded from the PACS (Carestream PACS; Carestream Healthcare) for CTA, while TOF sequences were downloaded for MRA. Post-contrast T1WI, T2WI and post-contrast MRA were downloaded when available.

### *Intermed IUHW Hospital, Ulaanbaatar, Mongolia*

With approval from the Intermed IUHW Hospital / International Medical Center Institutional Review Board in Ulaanbaatar, Mongolia, and with a waiver of informed consent, brain MRI studies with 3D-TOF MRA performed between 2014 and 2024 were retrieved from the institutional archive. These studies were reviewed to identify all patients admitted to the hospital during this period and were evaluated by a neuroradiologist. Patient anonymity was ensured through the use of de-identified data.

Additionally, a search of the institutional PACS (INFINITT version 3.0.11.5_B9P1) identified all brain MRI studies with 3D-TOF MRA performed between 2014 and 2024. The corresponding reports and images were reviewed by a consultant neuroradiologist and categorized as positive or negative for intracranial aneurysm. For positive cases, the size and location of each aneurysm were recorded. Axial T2, T1, 3D-TOF MRA, and post-contrast T1 sequences were manually exported from the PACS in DICOM format.

### *Koc University Hospital, Istanbul, Turkey*

A search was performed of the institutional PACS (Sectra IDS7; Sectra AB) for brain MRI and brain MRI angiography studies performed between 2014 and 2024. The corresponding reports were reviewed by a radiologist and categorized as positive or negative intracranial aneurysms. The level and location of each aneurysm were noted for the positive cases. CT angiography, TOF brain MR angiography, T2-weighted imaging (FSE, SPACE/CUBE, FIESTA/CISS), and post-contrast T1-weighted 3D images (e.g. SPGR, MPRAGE, BRAVO) images were manually exported from the PACS as DICOM images.

### *Liverpool Hospital, Liverpool, Australia*

A retrospective search was conducted within our institution’s PACS for CT angiography (CTA) of the brain. Search filters and interpretation codes were used to identify candidate studies, and each report was manually reviewed and categorised as positive or negative for intracranial aneurysm. Positive cases were further subdivided according to aneurysm location (e.g. internal carotid artery, middle cerebral artery, anterior communicating complex, posterior circulation). Inclusion criteria were age ≥18 years and imaging performed using a standard intracranial CTA protocol; studies with treated aneurysms, alternative vascular lesions or non-diagnostic scans were excluded. Axial CTA DICOM images were downloaded and anonymised using the RSNA Anonymiser tool, then subsequently uploaded to MD.ai for use in the study.

### *Memorial University of Newfoundland, St. John's, Newfoundland*

Memorial University of Newfoundland searched their institutional RIS using Nuance mPower (Nuance Communications) for cranial CT and MR angiograms between 2019 and 2024. Radiology reports were manually reviewed and categorized as positive or negative for intracranial aneurysms. Study images were then manually reviewed by a board-certified radiologist to confirm the absence/presence of aneurysm(s). Scans that did not meet the predetermined study inclusion criteria were excluded. Data pertinent to the study including the site of any aneurysm was collected. Pertinent patient demographic data was also collected according to the study protocol. Images were exported from the PACS (GE Healthcare Centricity) and were de-identified using RSNA Anonymizer software. Images were manually reviewed to ensure the absence of identifying health information and were then directly uploaded to the MD.ai platform for processing.

### *Philippine General Hospital, University of the Philippines, Manila, Philippines*

The conduct of this study in our institution was approved by the University of the Philippines Manila Research Ethics Board. Our radiology report database was searched for brain CT and MRI examinations that satisfy the study inclusion and exclusion criteria. Their images and relevant clinical information were manually reviewed by the study site investigators, and eligible patients were identified starting from the most recent examinations beginning in October 2024 and working backwards until a sufficient cohort was reached. The required imaging data were then extracted into DICOM format, and all identifying information were removed. The anonymized studies were reviewed by an experienced neuroradiologist to identify the presence or absence of cerebral saccular aneurysm. Positive cases were also assessed for aneurysm location and signs of rupture. The anonymized examinations were uploaded to the MD.ai online platform for collation and processing by the RSNA AI Challenge Team.

### *Prince Sultan Military Medical City, Riyadh, Saudi Arabia*

The Institutional Review Board at Prince Sultan Military Medical City approved this retrospective study and waived informed consent. The keyword (aneurysm) was searched within radiology reports using the local Pictures Archiving and Communication System (GE Healthcare's Centricity Universal Viewer and Nuance PowerScribe 360), from January 1, 2011, through December 31, 2024. We manually reviewed and analyzed all CT and MR angiograms through the head to ensure adherence to the predefined criteria. Data including patient age, sex, race, ethnicity, anatomical location of the aneurysm, and the date of study were collected. DICOM-formatted de-identified images were downloaded from the PACS. The DICOM files were then uploaded directly to MD.ai for final processing.

### *Queen's University, Kingston, Ontario, Canada*

The KHSC–Queen's site investigator conducted a search using Nuance mPower (Nuance Communications) for MRA and CTA of the brain performed between January 1, 2022, and December 1, 2024. The search was restricted to outpatient status and patients aged 18 years or older. For positive cases, the following keywords were included: aneurysm, saccular, intracranial aneurysm. For negative cases, the keywords were: pseudoaneurysm, mycotic, AVM, and no intracranial aneurysm. Each examination was manually reviewed to confirm that inclusion criteria were met, and cases were excluded according to the exclusion criteria. A total of 198 cases were selected. For each positive case, the aneurysm location was recorded. For each negative case, a second review confirmed that no aneurysms were visualized. Patient age and sex at birth were also documented. All selected MR studies were downloaded in anonymized form from PACS (Universal Viewer – GE) to a local workstation within the hospital network using RSNA Anonymizer (Radiological Society of North America). The axial T2, TOF MRA, and CTA source series were manually identified within each study and uploaded to MD.ai for annotation.

### *University Hospitals Cleveland Medical Center, OH, USA*

This retrospective study received institutional approval with a waiver for informed consent. The EMR (EPIC) and PACS systems (SECTRA) of University Hospitals Cleveland Medical Center were manually queried by the radiology informatics team to identify potential cases applicable for this study using the date range from January 1, 2020, through January 1, 2024, and applicable exam codes. The final reports for these exams were treated as the gold standard and were manually reviewed by radiologists to collect the requested information. This extracted report information was paired back with the requested patient demographic data in Excel spreadsheets. The exam imaging sequences were de-identified using the RSNA Anonymizer. The accompanying spreadsheet was updated to contain anonymized keys for patient, MRA exam, and MRI exam fields. Data was uploaded per instruction.

### *University of California Irvine, Irvine, CA, USA*

A retrospective query was performed to identify all head CT angiograms (CTA) acquired across emergency, inpatient, and outpatient settings between January 1, 2024, and December 31, 2024. A language model (Llama3.3-70B) was used to parse each radiology report for the presence of at least one saccular intracranial aneurysm measuring >= 3 mm as well as determine aneurysm location. Subsequently, an expert neuroradiologist manually verified every examination flagged as positive by the model. Patient metadata, including age and gender, were aggregated for the final cohort.

Study enrollment followed a hierarchical prioritization scheme to maximize clinical utility. For verified positive cases, selection first prioritized patients with at least one prior MRI examination (T1-post contrast, T2, or Time-of-Flight MRA), followed by aneurysms in rare locations (posterior circulation). The remaining positive cohort was filled through random selection until enrollment targets were met. Negative control cases were limited to patients with both the index CTA and corresponding prior MRI imaging. All imaging data was deidentified using custom in-house Python software, including voxel-level and DICOM header anonymization (remapping unique identifiers, clipping upper bounds of patient age to a limit of 89 years, shifting temporal data, and removing all other sensitive fields).

### *University of California San Francisco, San Francisco, CA, USA*

With approval of the UCSF institutional review board, radiology reports from UCSF examinations performed between 2008 and 2024 were queried for the term “aneurysm,” restricted to CTA and MRA/MRI studies. Reports were exported from the institutional radiology information system and manually reviewed by a neuroradiology fellow and a board-certified radiologist. Reports were classified as positive or negative for aneurysm. Negative examinations included studies in patients with previously treated aneurysms without residual or recurrent aneurysm. For positive examinations, details including patient demographics, aneurysm size, aneurysm location, rupture status (ruptured vs unruptured), and whether the aneurysm was treated or not was recorded. Positive studies and an equal number of negative studies were then downloaded from the institutional PACS and de-identified. Each study was manually reviewed to confirm diagnostic image quality and protocol appropriateness (including thin-section reconstructions when available for CTA and MRA/MRI) and to ensure complete removal of any protected health information, including burned-in annotations or screen-capture content.

### *University of Health Sciences, Umraniye Training and Research Hospital, Istanbul, Turkey*

A retrospective search was conducted in the institutional PACS (RAD-X Simplex) to identify intracranial vascular imaging studies performed between 2020 and 2025. Both CT angiography (CTA) and MR angiography (MRA) examinations were included. The search strategy incorporated the keyword “saccular aneurysm.” All radiology reports were manually reviewed by a radiologist with 6 years of experience, and studies were classified as positive or negative for intracranial saccular aneurysm. For positive cases, the aneurysm location and the presence of multiple aneurysms were recorded. Exclusion criteria included patients younger than 18 years of age and examinations with significant motion artifacts resulting in nondiagnostic image quality. For CTA, only studies with a slice thickness of ≤1.0 mm were included. For MRA, time-of-flight (TOF) sequences were selected. All imaging studies were de-identified using the RSNA DICOM Anonymizer.

### *University of Utah, Salt Lake City, Utah, USA*

The primary site investigator searched their institutional RIS (Radiant; Epic Systems Corporation) using Nuance mPower (Nuance Communications) with the following terms and limiting the examination to MRA head and CTA head modalities: “PICA aneurysm,” “AICA aneurysm,” “SCA aneurysm,” “basilar artery aneurysm,” “basilar tip aneurysm,” “PCA aneurysm,” “ICA aneurysm,” “ACA aneurysm,” and “MCA aneurysm.” The date range included examinations from 7/6/2006 through 2/6/2025. All of the following sequences were collected, if available within a 7-day acquisition window: MRA, T2, T1 post-contrast, MRA post-contrast, MRA time-of-flight, and CTA. Each examination was manually reviewed to confirm that inclusion criteria were met, and cases were excluded according to the exclusion criteria.

For each case, basic demographic information including sex at birth, age at the time of exam, race, and ethnicity were recorded. Details regarding the aneurysm(s) present, including size, location, active rupture status, and treatment status, were also recorded. An approximately equal number of normal control patients were collected by searching for MRAs and CTAs using CPT codes. An effort was made to obtain the normal comparisons in a similar distribution of exam dates. We also made an effort to collect a roughly similar number of aneurysms from each of the designated aneurysm locations.

These studies were manually exported from the PACS (IntelliSpace PACS; Philips Healthcare) to a local workstation within the hospital network. The studies were anonymized and then uploaded to MD.ai for annotation.

# RSNA Brain Aneurysm AI Challenge

## Annotation Instructions

**2/3/2025**

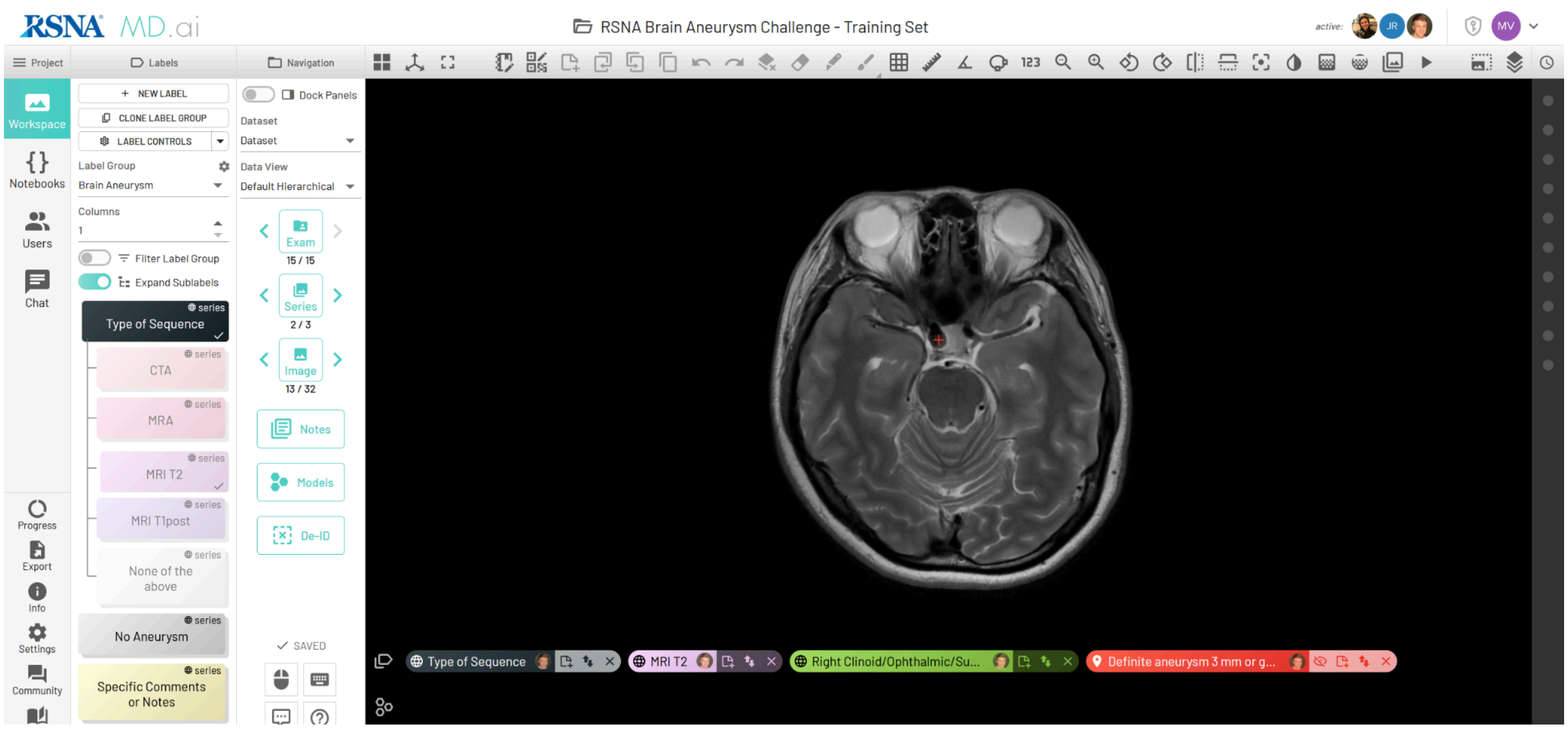

## TABLE OF CONTENTS

Radiological Society of North America

# 1. MD.ai Login and Setup for Annotation

- ## Logging in to RSNA MD.ai
  - Navigate to the following link: [https://rsna.md.ai](https://rsna.md.ai)
  - Click the **LOG IN** button in the upper right corner and enter your credentials:

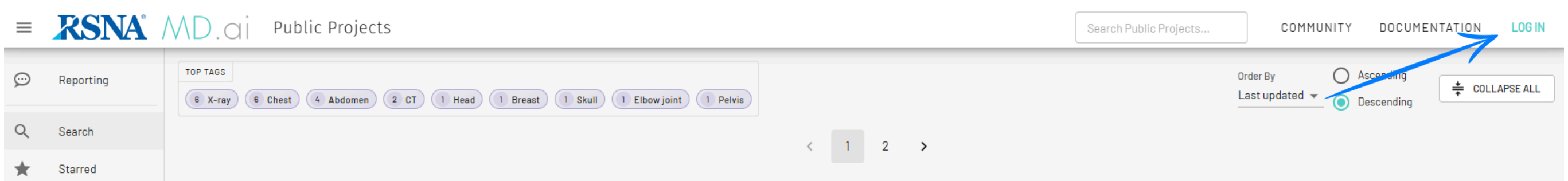

- ## Project Selection and Launch
  - After logging in, select **My Projects** from the left panel:

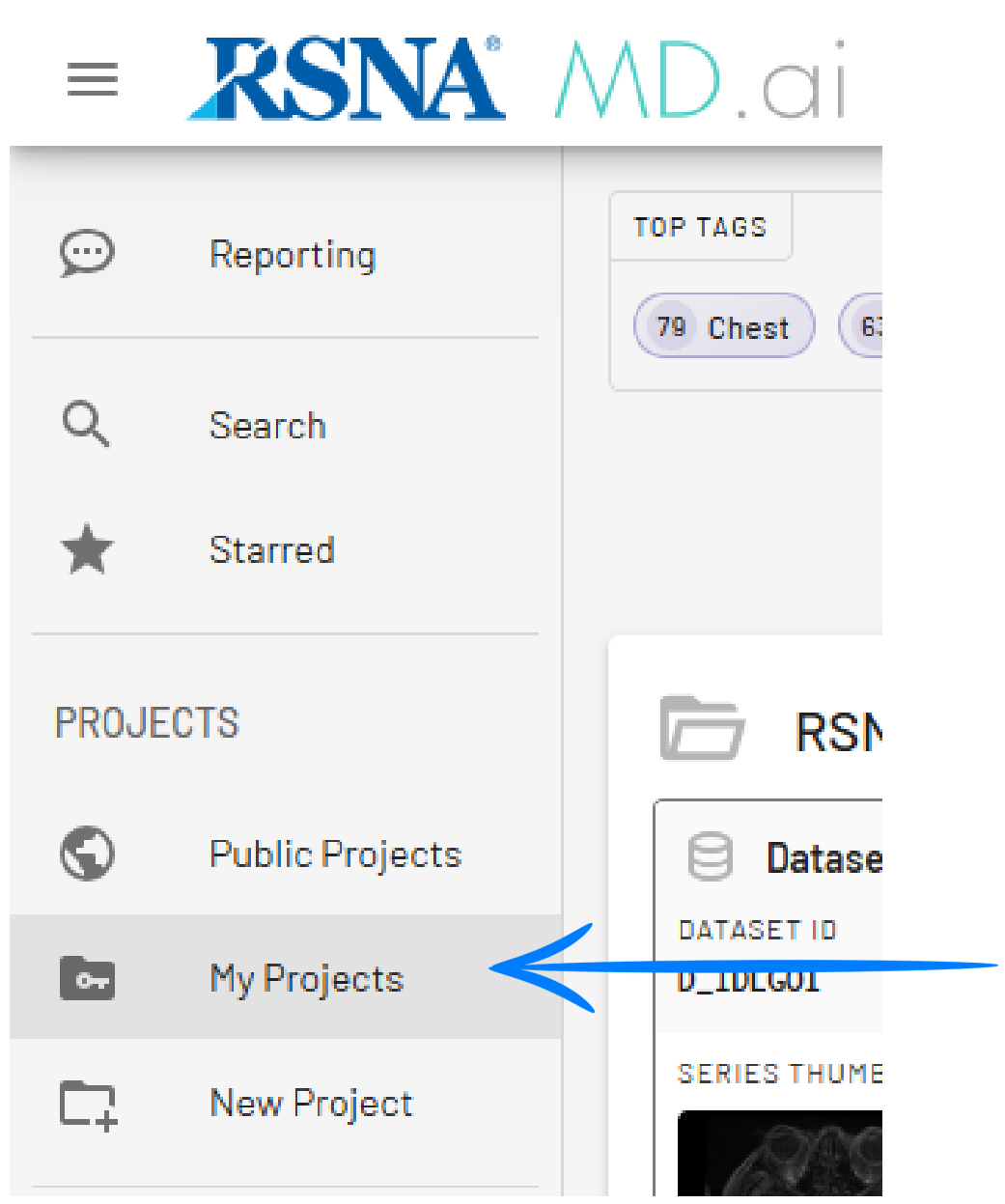

- Click **Launch** for the RSNA Brain Aneurysm AI Challenge project:

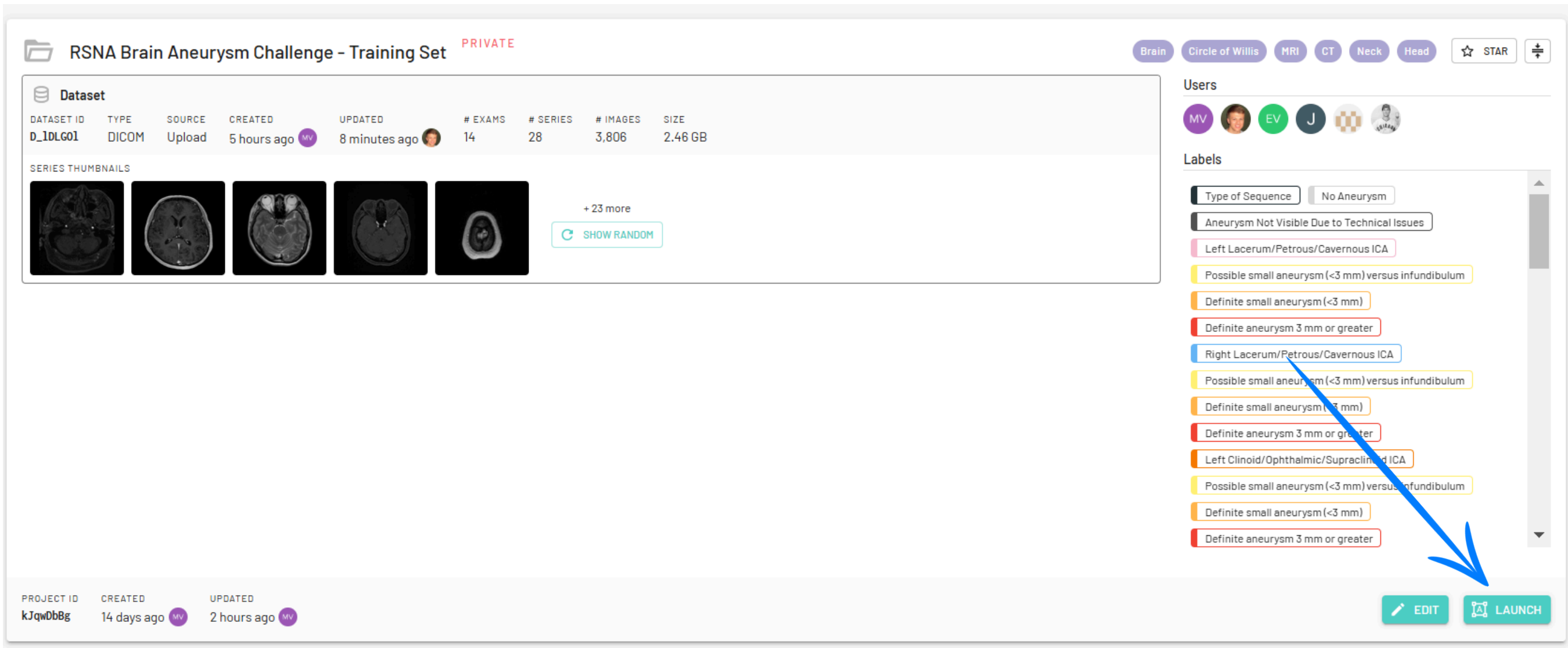

- Once launched, you will see the annotation workspace. The project pane will open with the first case that you need to annotate:

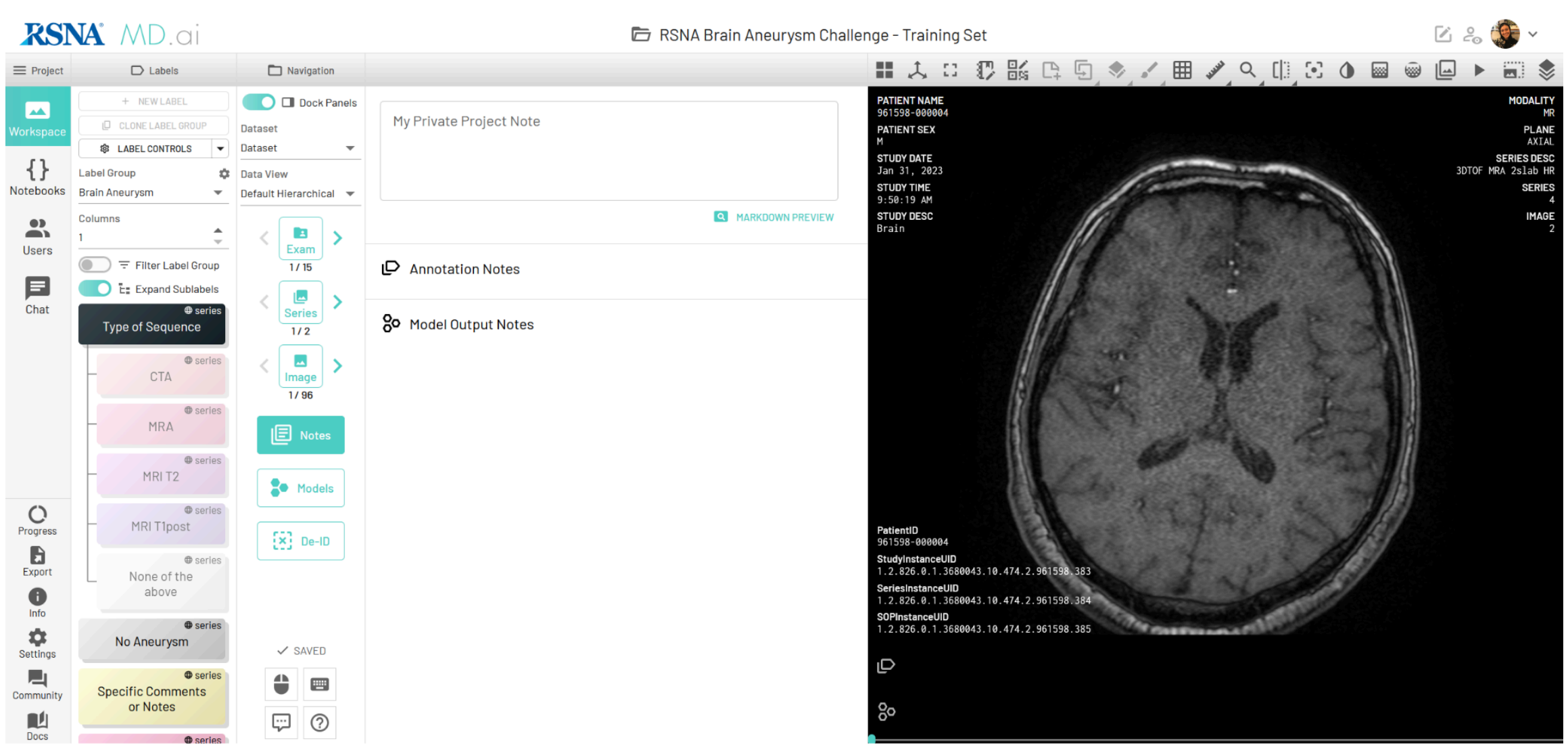

# 2. Annotation Process

- **Label Group Selection**
  - Click the down caret under **Label Group** and make sure that the label group **Brain Aneurysm** is selected:

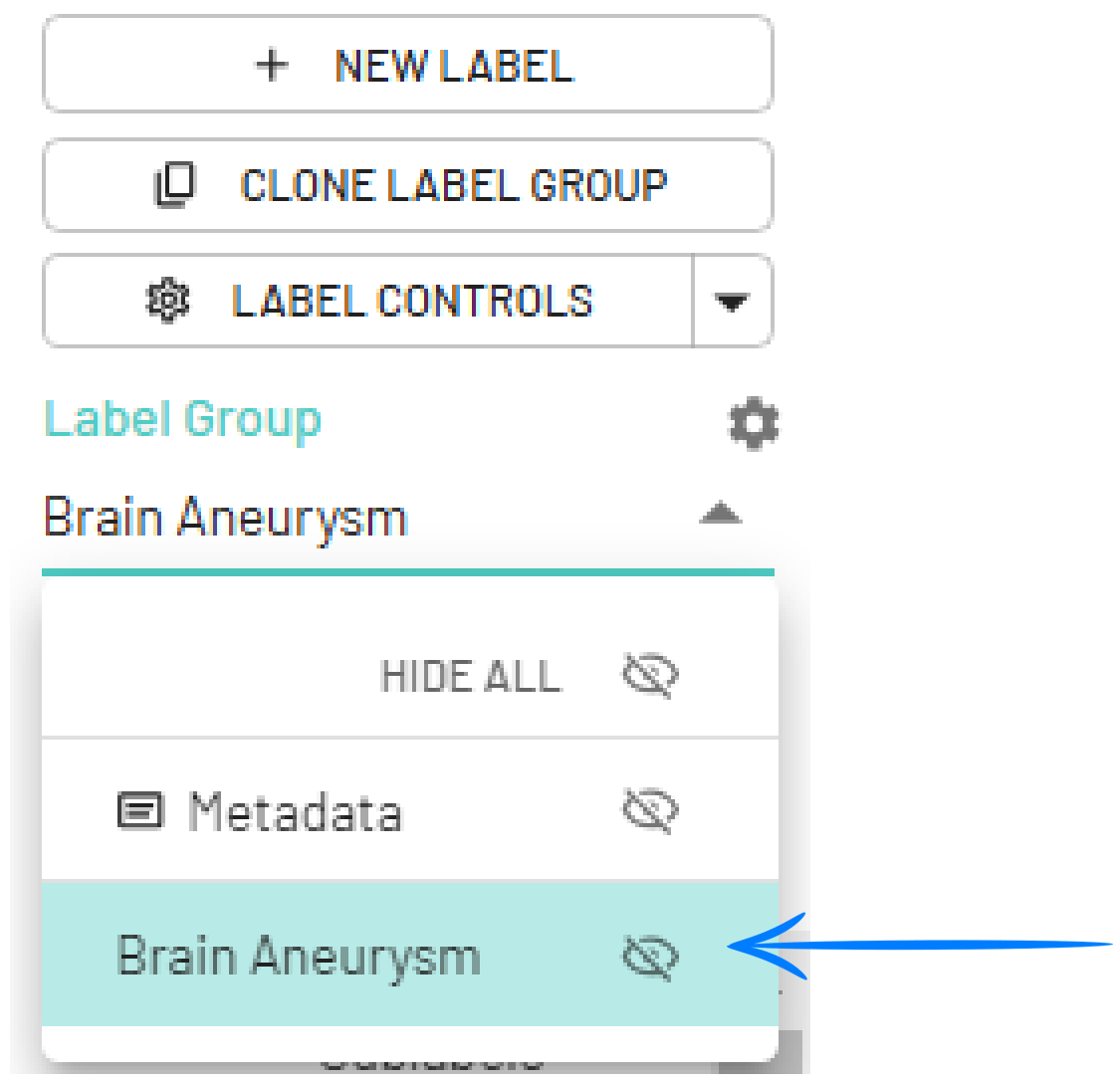

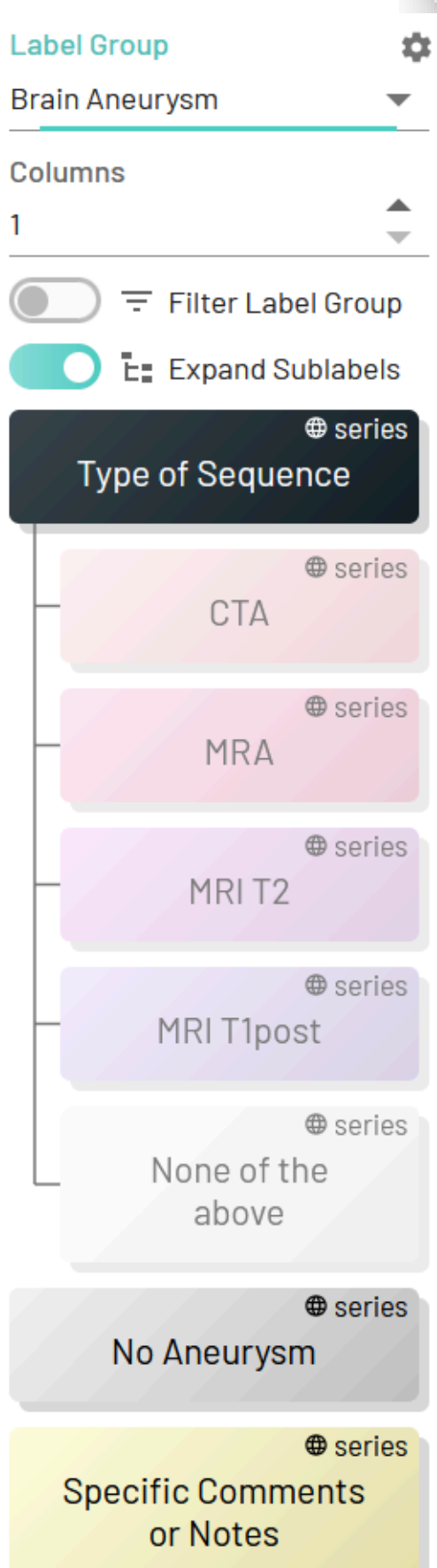

The buttons in the left column are the labels that you will apply to each series. Select **Expand Sublabels** to view all labels. Use the scrollbar to scroll through all available labels.

Radiological Society of North America

- ## Navigate to Cases
    - You can select the exam, series, or specific image to work on using the Navigation column:

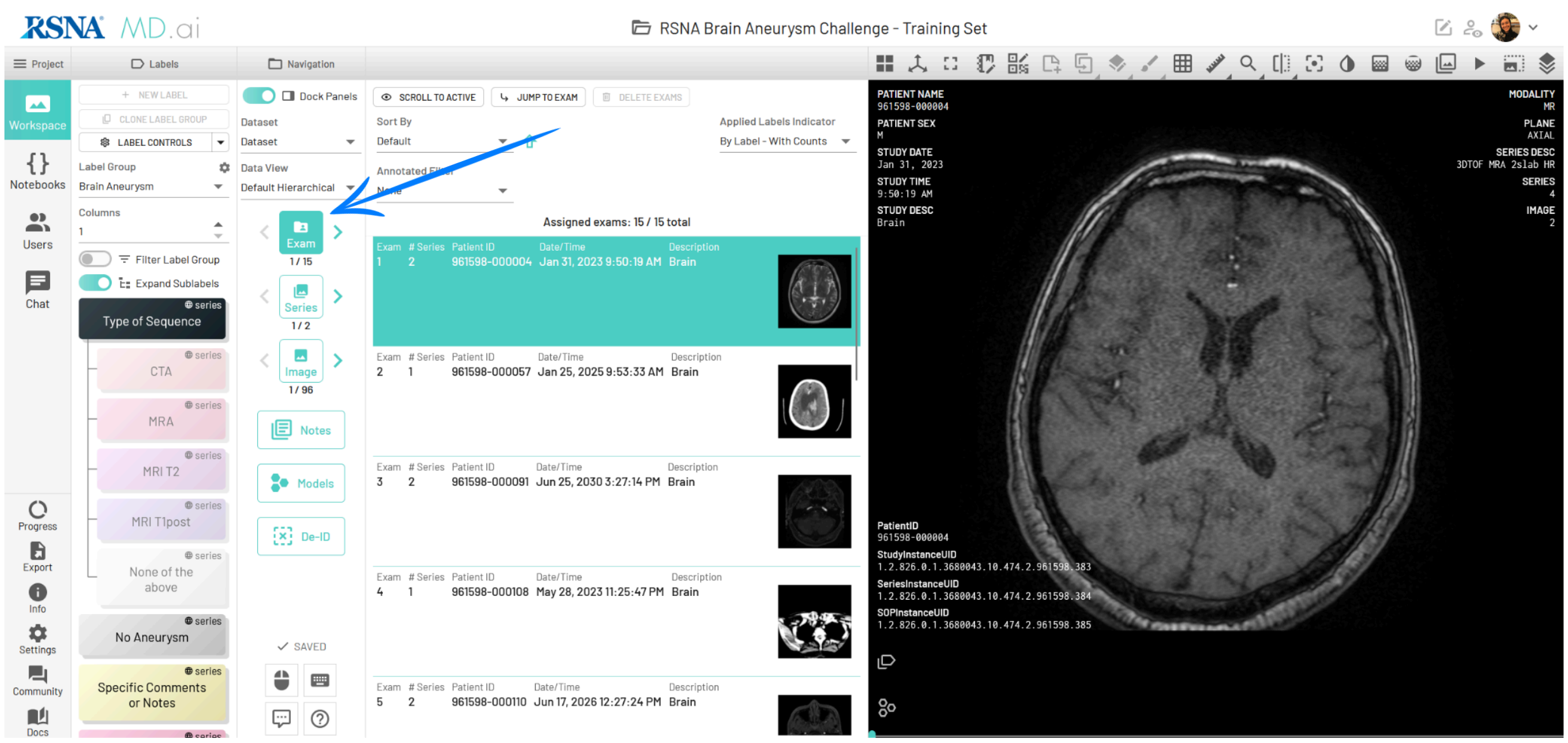

- You can view the cases at the exam level, series level, or image level by selecting the corresponding icon and navigating using arrows:

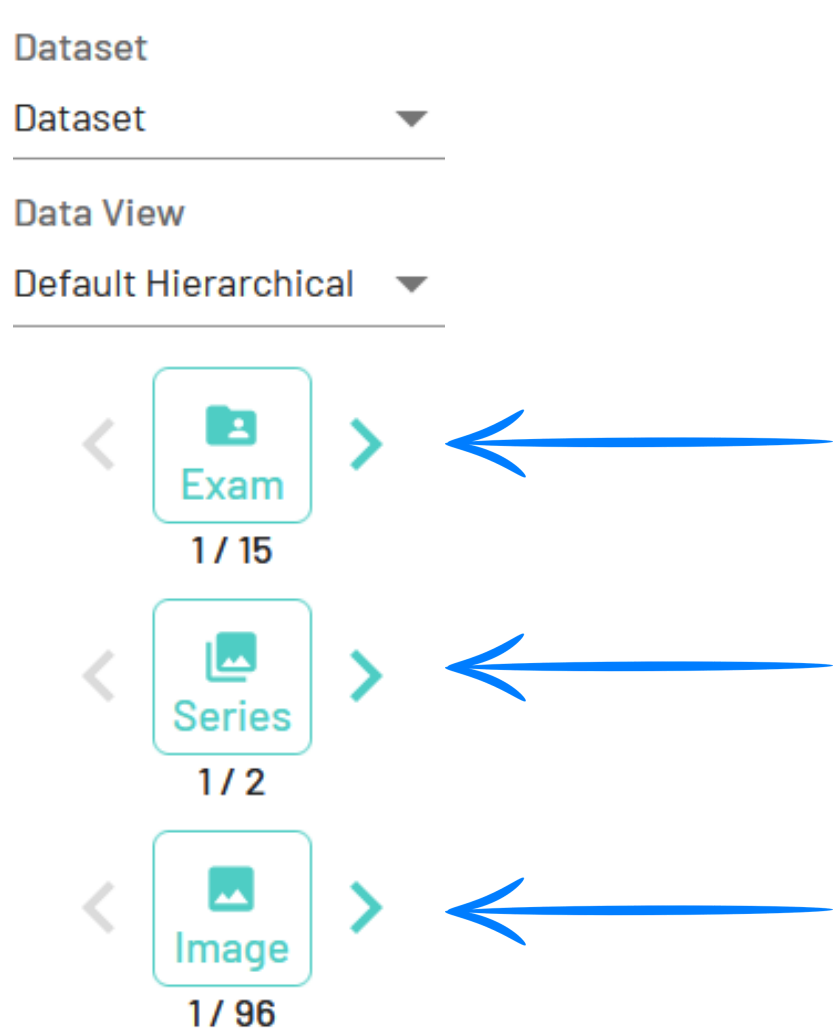

## Using 3D Multi Planar Reconstruction (MPR)

If you would like images to be displayed in other orientations, you can enable the 3D Multi Planar Reconstruction (MPR) feature. Activating MPR allows you to view the image in multiple planes on one screen. The feature overlays reference lines and crosshair pointers on other planes while you work on a specific view. Locate the MPR function on the top left corner and select your preferred layout:

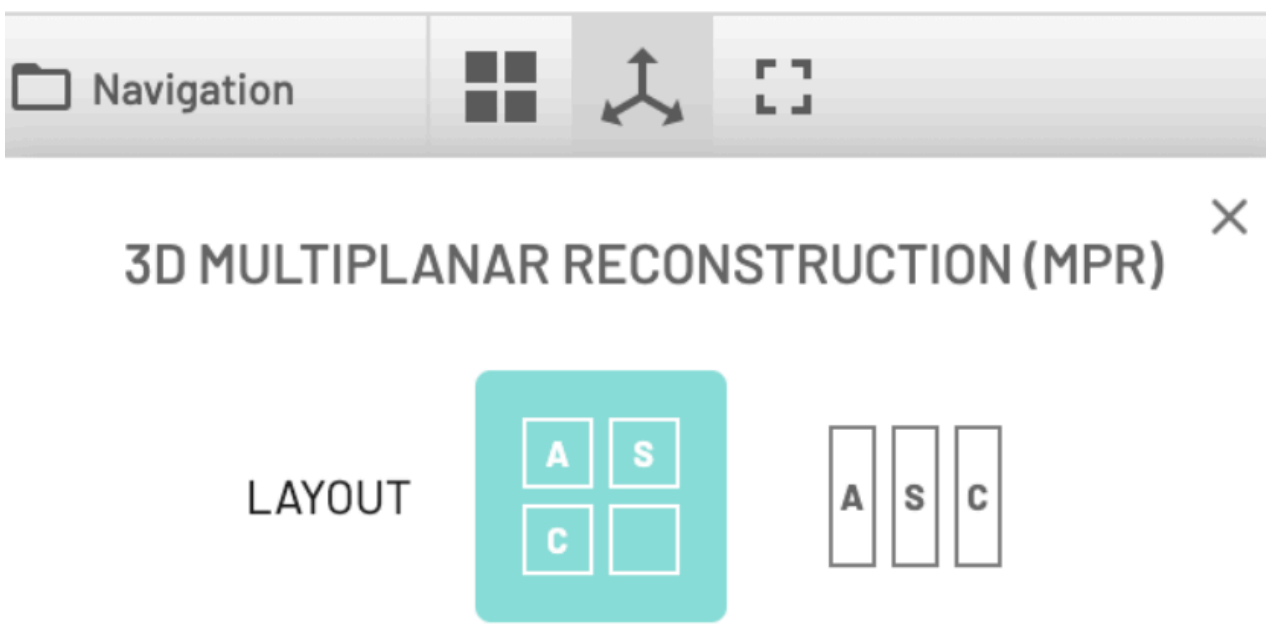

Additional viewing options including **Maximum Intensity Projection (MIP), Minimum Intensity Projection (MinIP), and Average** can be selected while a reconstruction is active by selecting the MPR icon again.

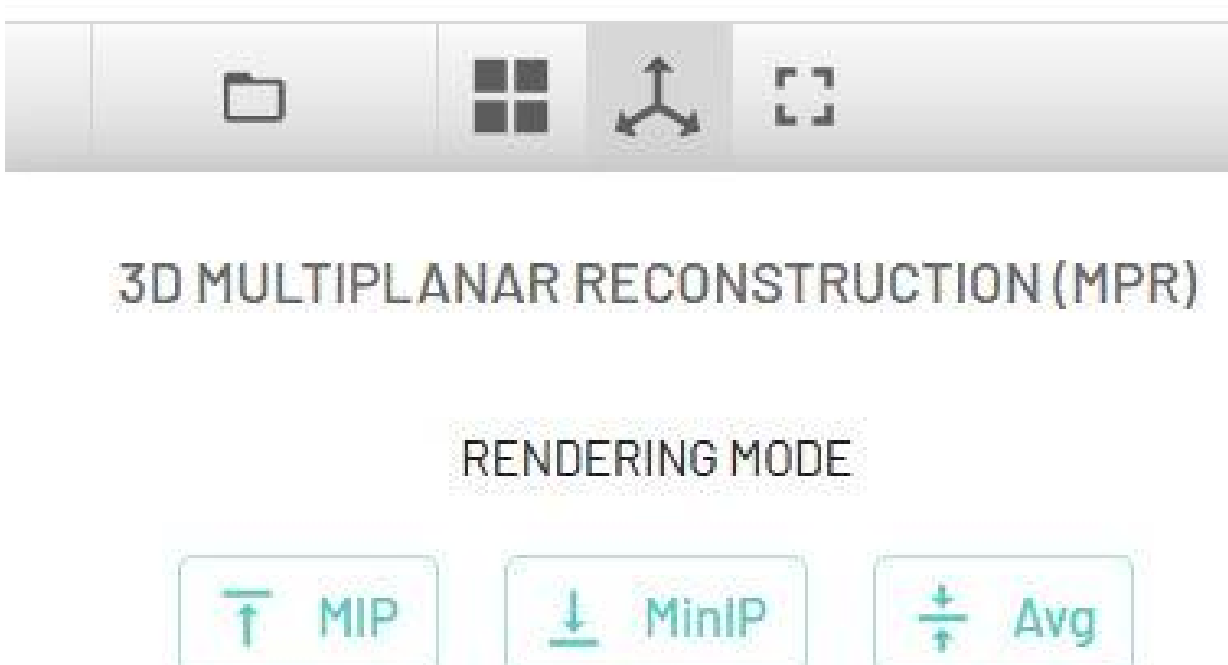

## Measuring Using the Ruler Tool

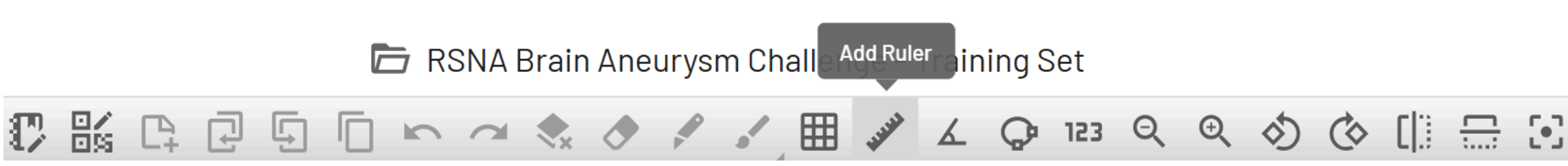

You can measure aneurysms using the **Ruler tool** found in the top toolbar. Using the ruler tool, follow these steps:

- Click the ruler icon in the toolbar to activate the measurement tool.

- Mark the first point by clicking at one edge of the aneurysm where you want to start measuring.
- Move your cursor to the opposite edge of the aneurysm and click again to complete the measurement.
- Check the measurement. The tool will display the distance in millimeters.
- If adjustment is needed, slick and drag the points to refine the measurement.
- To remove it, hover your mouse over the measurement and press delete (X).

## Other Basic MD.ai functions

- You can scroll through each image stack using: (1) the mouse wheel, (2) left-right arrow keys, (3) a slider at the bottom of each image and (4) using the cine arrow button ▶.
- Left mouse click and drag, left-right/up-down will change window width and level.
- Push down center mouse wheel and drag will pan image.
- Notice that a blue reference line will move with the image on the right pane for localization.
- Click the following icons in the bottom left navigation pane to view the mouse and keyboard shortcuts.

# 3. Annotation Instructions:

Annotations should be made following these below steps (also see video demo):

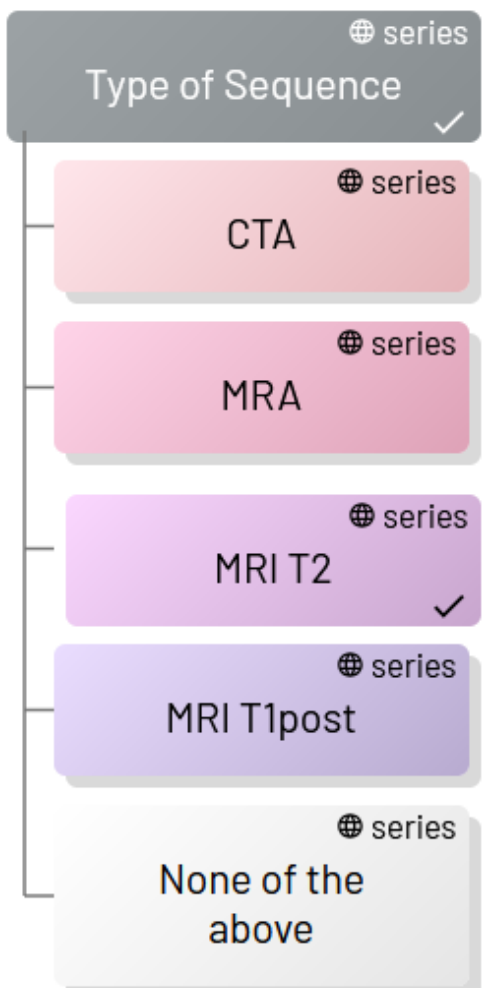

1. Choose the appropriate imaging modality for the series by selecting the **Type of Sequence** parent label and choosing one of the following sublabels:

   - **CTA**
   - **MRA**
   - **MRI T2**
   - **MRI T1 Post**
   - **None of the Above**

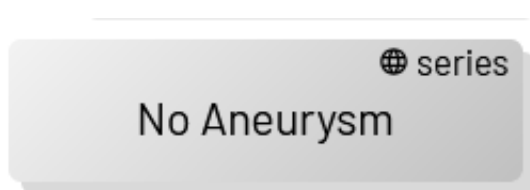

2. If no aneurysm is present in the series, select **No Aneurysm**.

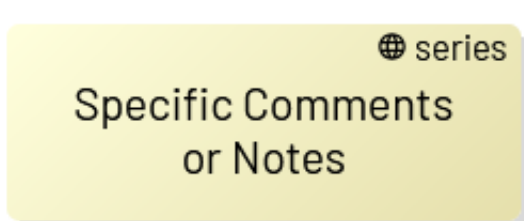

3. If you have specific comments you want to provide about a case, you can select the **Specific Comments or Notes** label and write free text.

4. If there is an aneurysm in this case, select the aneurysm location. Only include intracranial saccular aneurysms. Exclude fusiform, mycotic, and pseudoaneurysms, as well as aneurysms associated with AVMs, unless a flow-related aneurysm is located outside the nidus (please note any of these findings in the **Specific Comments or Notes** label). The classification of the aneurysm location should be based on the vessel of origin. For aneurysms located at the site of a side branch, classify the aneurysm as originating from the side branch artery (e.g., an aneurysm at the ICA/PCom junction should be labeled as PCom). There will be some ambiguity in aneurysm location, so please select whatever you believe to be the best choice of the available options:

   - **Left Lacerum/Petrous/Cavernous ICA**
   - **Right Lacerum/Petrous/Cavernous ICA**
   - **Left Clinoid/Ophthalmic/Supraclinoid ICA**
   - **Right Clinoid/Ophthalmic/Supraclinoid ICA**
   - **Left MCA (any segment)**

- **Right MCA (any segment)**
- **Anterior Communicating Artery**
- **Left ACA (any segment)**
- **Right ACA (any segment)**
- **Left Posterior Communicating Artery**
- **Right Posterior Communicating Artery**
- **Left PCA (any segment)**
- **Right PCA (any segment)**
- **Basilar tip**
- **Other Posterior Circulation (e.g., mid-basilar, vertebral, PICA, AICA, SCA)**

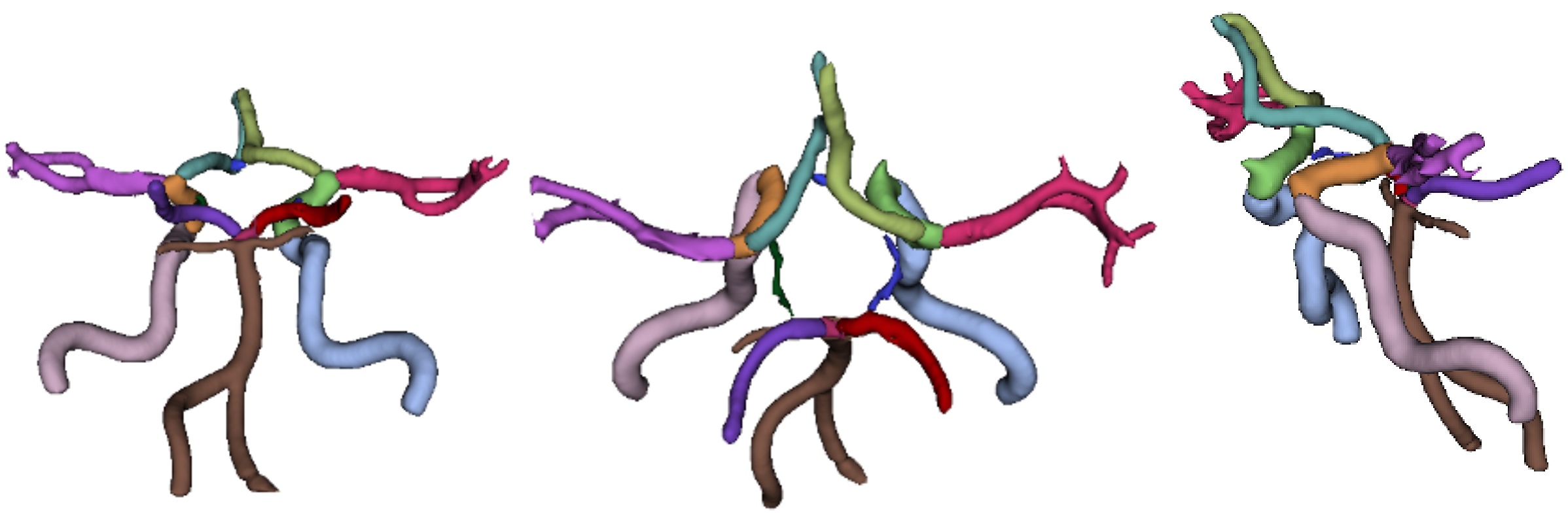

5. For each location label, also select one of the following sublabels:

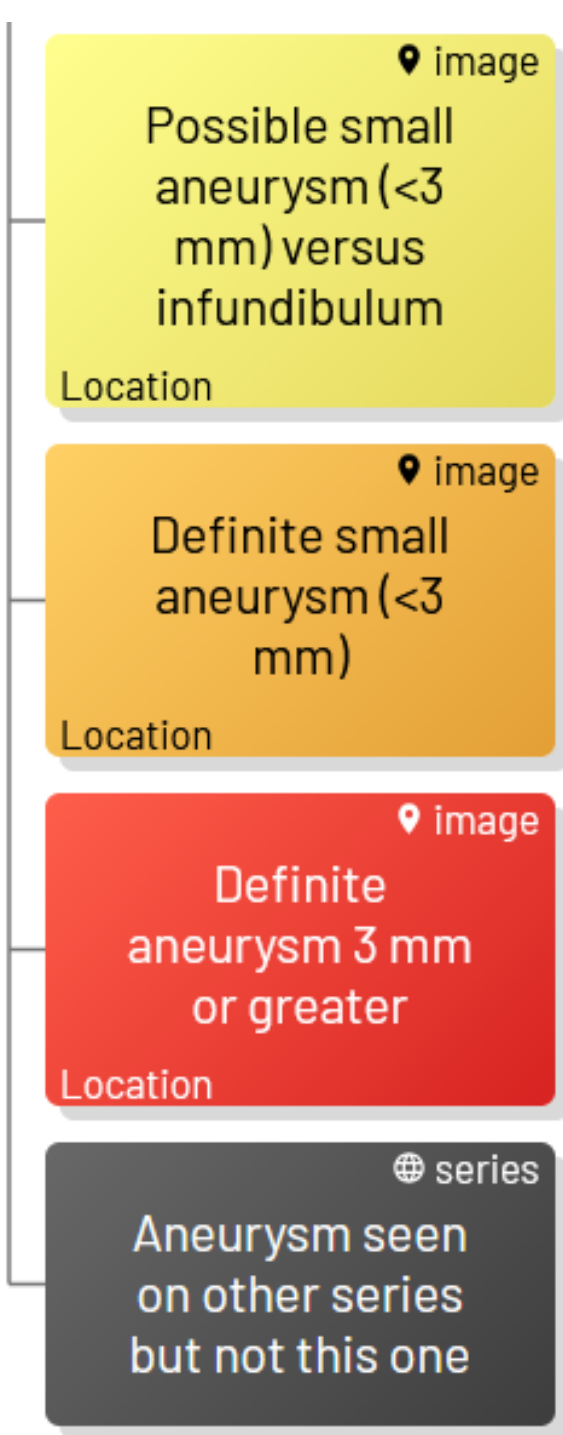

**- Possible small aneurysm (<3mm) versus infundibulum:** If you are unsure about the presence of the aneurysm you can select this label. You can use the measuring tool if you are uncertain of size. These cases will be reviewed by additional neurointerventional annotators.

**- Definite small aneurysm (<3mm)**

**- Definite aneurysm 3 mm or greater**

**- Aneurysm seen on other series but not this one:** If there is an aneurysm seen on the angiographic sequence

(CTA/MRA; look at these first if out of order) but it is not visible on the MRI sequence (T2/T1post) due to technical limitations (e.g., poor resolution, artifact), select this label.

6. After you select one of the above options (excluding the last label, **Aneurysm seen on other series but not this one)**, then click on the location of the aneurysm in the image to save a localizer annotation.

7. Keep in mind that cases can have more than one aneurysm. In cases with more than one aneurysm, you should select each aneurysm location and size separately.

8. Include only intracranial saccular aneurysms. Do not include cervical ICA Aneurysms, fusiform aneurysms, mycotic aneurysms or pseudoaneurysms.

9. If the aneurysm has been treated or is partially thrombosed, the aneurysm should be marked as being present only if there is residual/recurrent aneurysm. If there is no visible residual/recurrent aneurysm, mark it as **No Aneurysm**. For size please measure the residual/non thrombosed portion of the aneurysm.

Basic workflow:

1. Send your IP to Evan Calabrese. Note that this can change based on your network, VPN, and even can change over time in some instances. If you need to send multiple that is ok. Whatsmyip.com is a good resource.

2. Download the latest stable version of 3DSlicer: https://download.slicer.org/

3. Once installed, open slicer and install the monai label plugin: https://docs.monai.io/projects/label/en/latest/quickstart.html#install-monai-label-plugin-in-3d-slicer

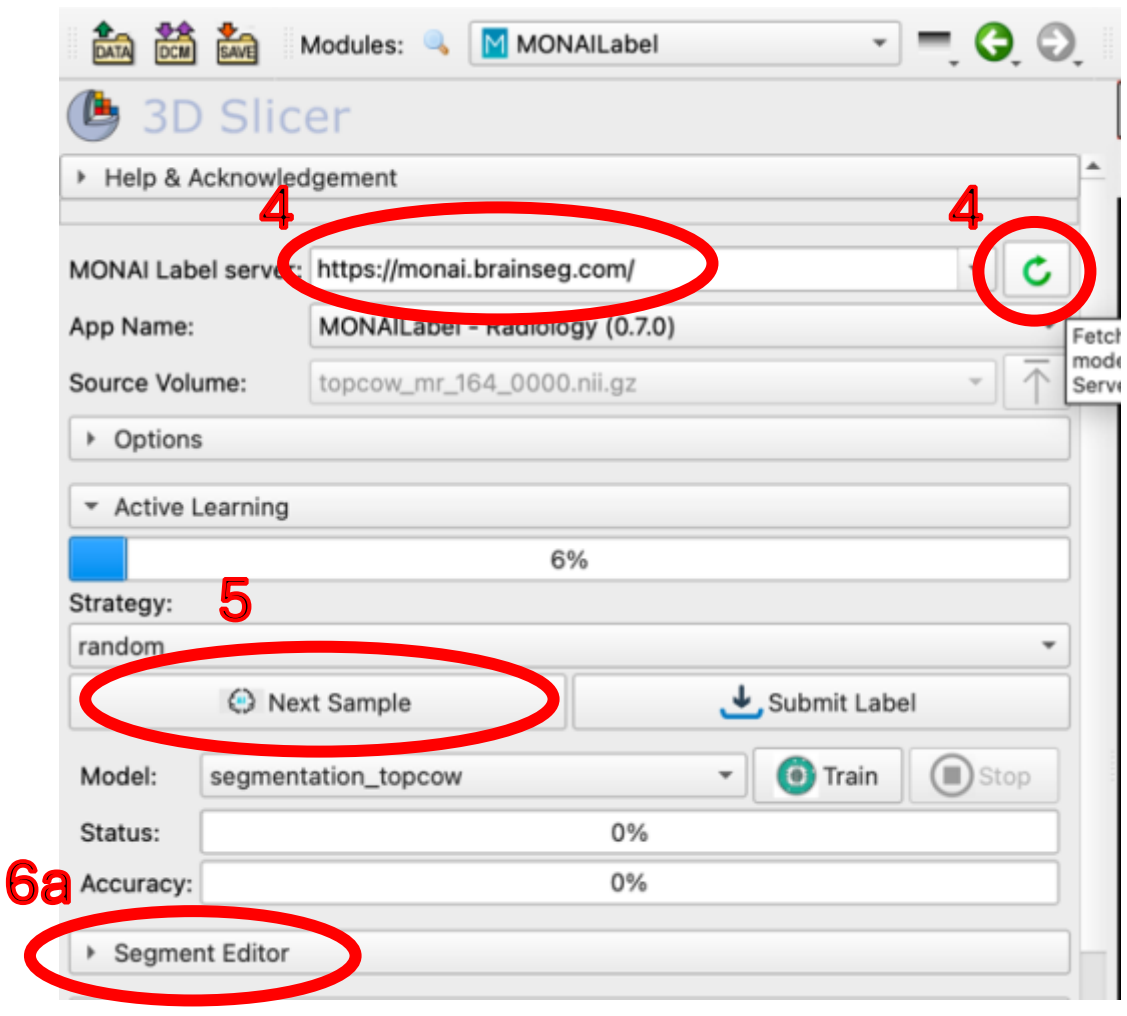

4. Open the monai label plugin in slicer. In the "MONAI Label Server" field, enter https://monai.brainseg.com Note that https is required. Click the green circle arrow button to connect.

5. Click the "next sample" button to get a random case from the server.

6. Do the annotations.
   a. Click the "segment editor" bar to expand the segment editor tools.

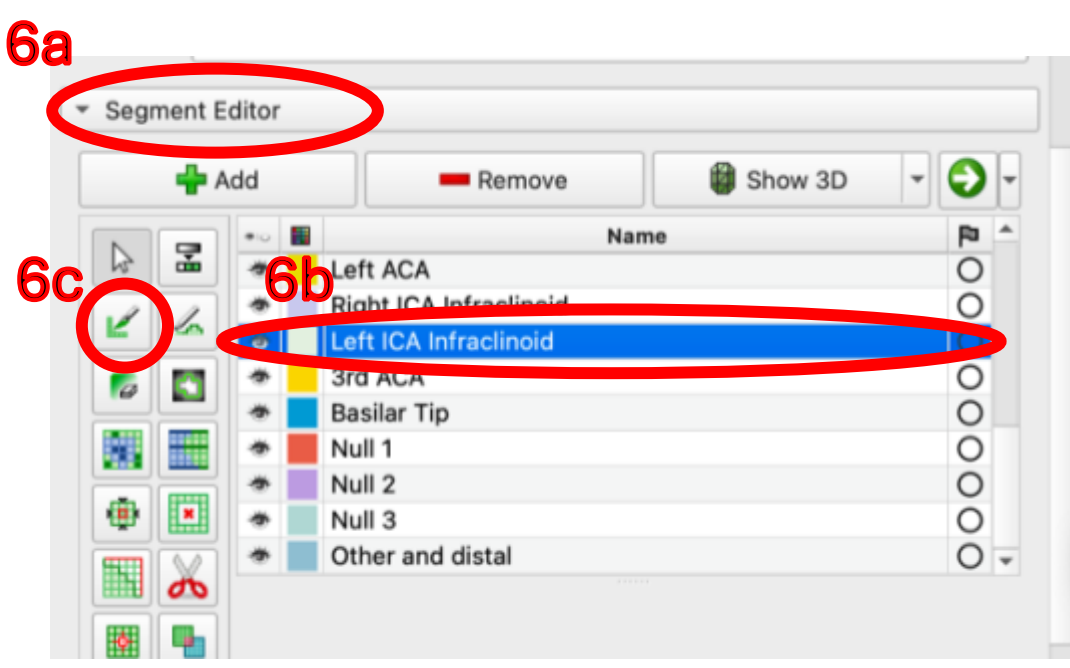

   b. Select the label you want to edit by clicking on the description (it will highlight).

   c. Once you have a label selected, select the "paint" tool (buttons along the left side of panel).

   d. Paint tool options will appear below the list of labels.

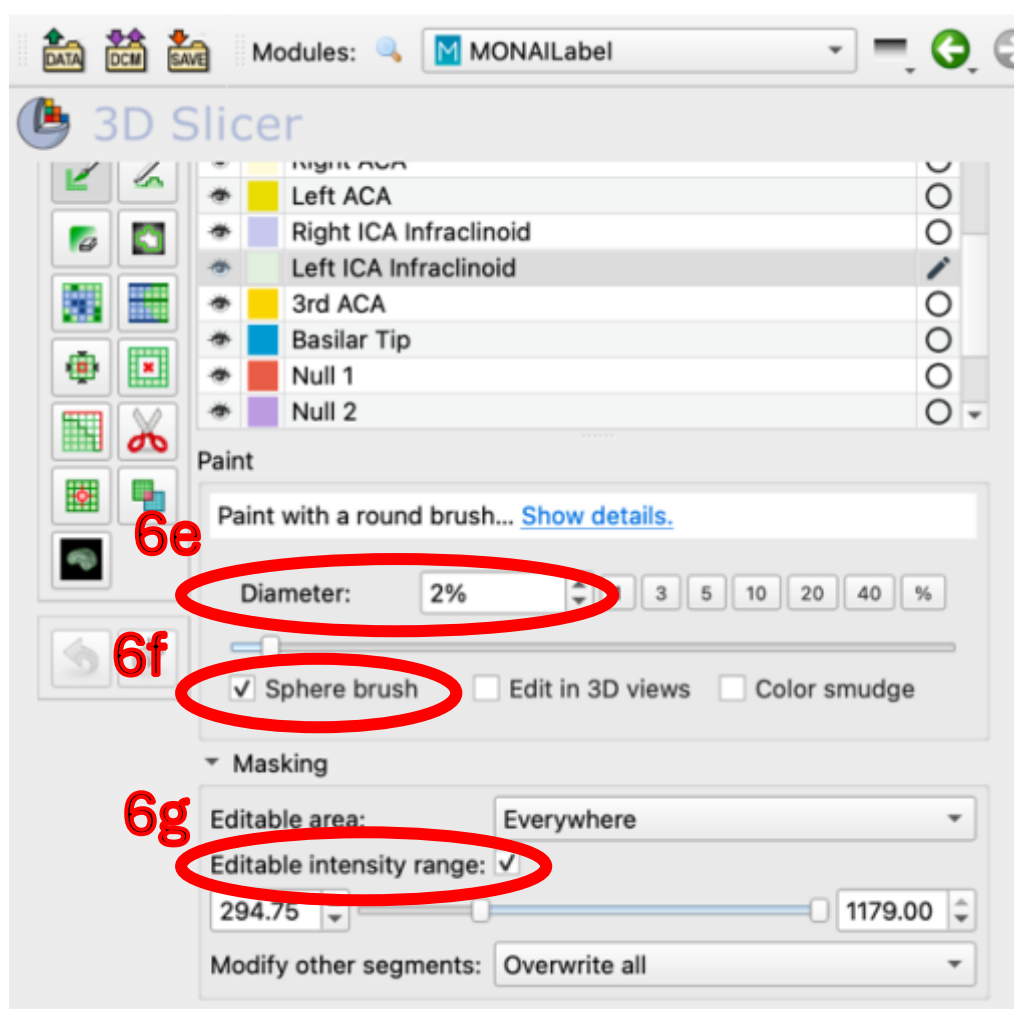

   e. Use the "diameter" slider to change the brush size.

   f. Check "sphere brush" to edit in a sphere rather than slice by slice.

   g. Check "editable intensity range" to edit within a specific range of intensities, which you will have to specify (hint: placing the cursor over the image shows the local intensity value in the bottom left "data probe" panel").

h. Select "editable area" which is the area that you can paint on.
   - CTAs should have all major unsegmented vessels already labels as "Other and Distal" label, which is meant to make it easier for you (just set editable area to inside other/dista).
   - MRAs do not have distal vessels segmented, but you can use intensity range to easily segment vessels. Make sure editable area is "everywhere".

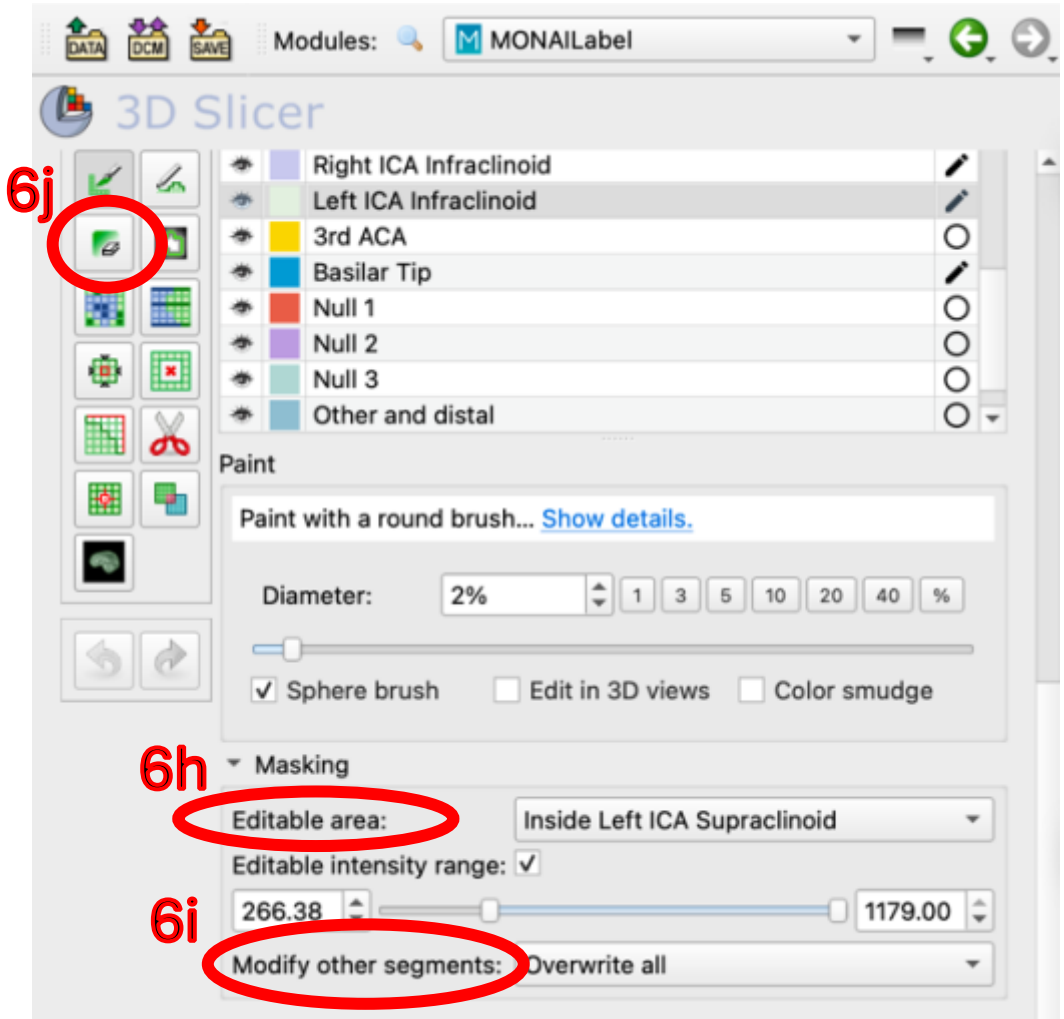

i. Select "modify other segments" to choose which (if any) segments you can write over.

j. If you need to erase a segmentation select "eraser" tool.

k. If you want to edit a different label, just click a different one from the list.

7. Click the "submit label" button to upload your work back to the server. Note that its not set up to save your work and come back later. You need to fully complete a case during your session. You can always leave the application open and come back later if you need to step away during a case.

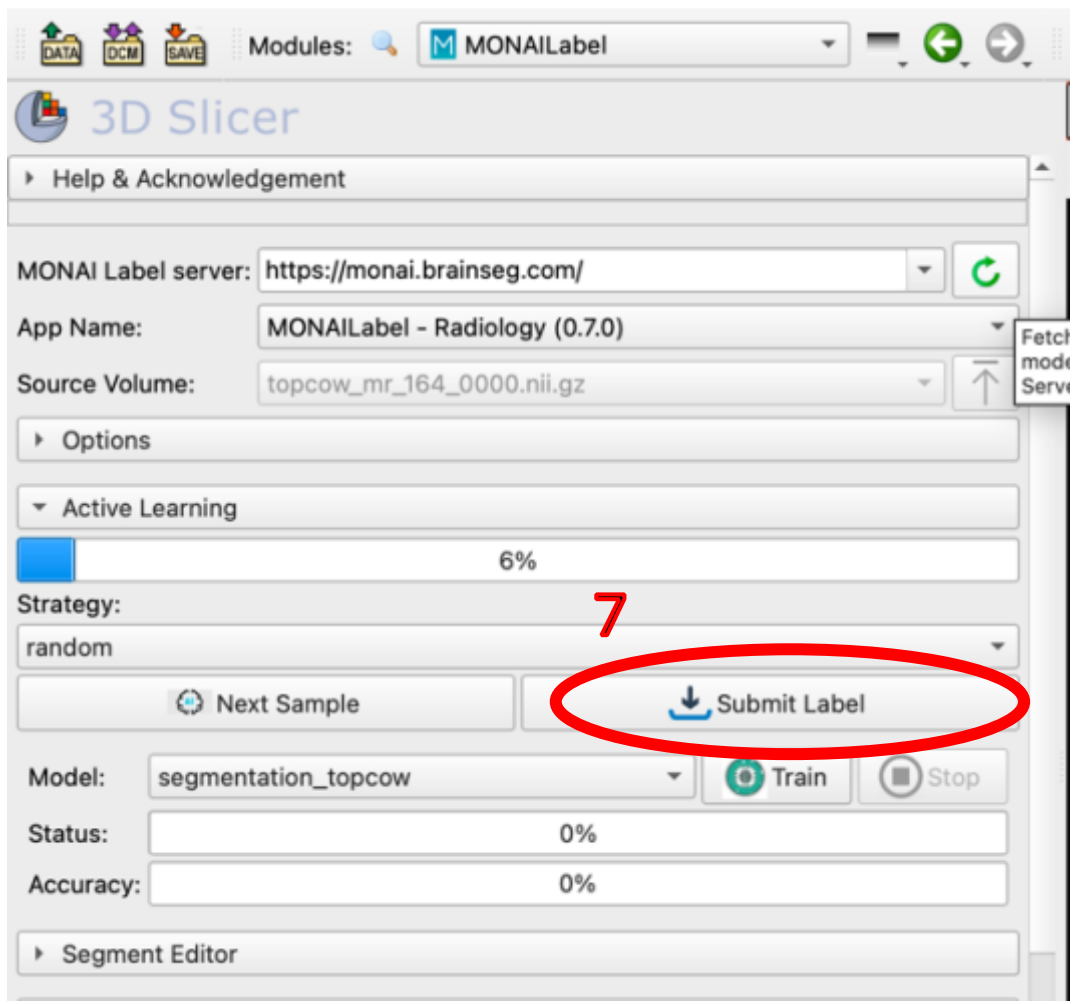

Annotation:
Only the part of the artery with flow (contrast-filled or with hyperintense MR signal) should be included, it is okay to skip a segment.

Steps:

1. Add infraclinoid ICAs to skull base to include petrous segment (best seen on sag)

2. Extend infraclinoid ICAs to anterior genu of cavernous carotid (also on sag).
   Note: this will require overwriting some of the existing supraclinoid ICA label.

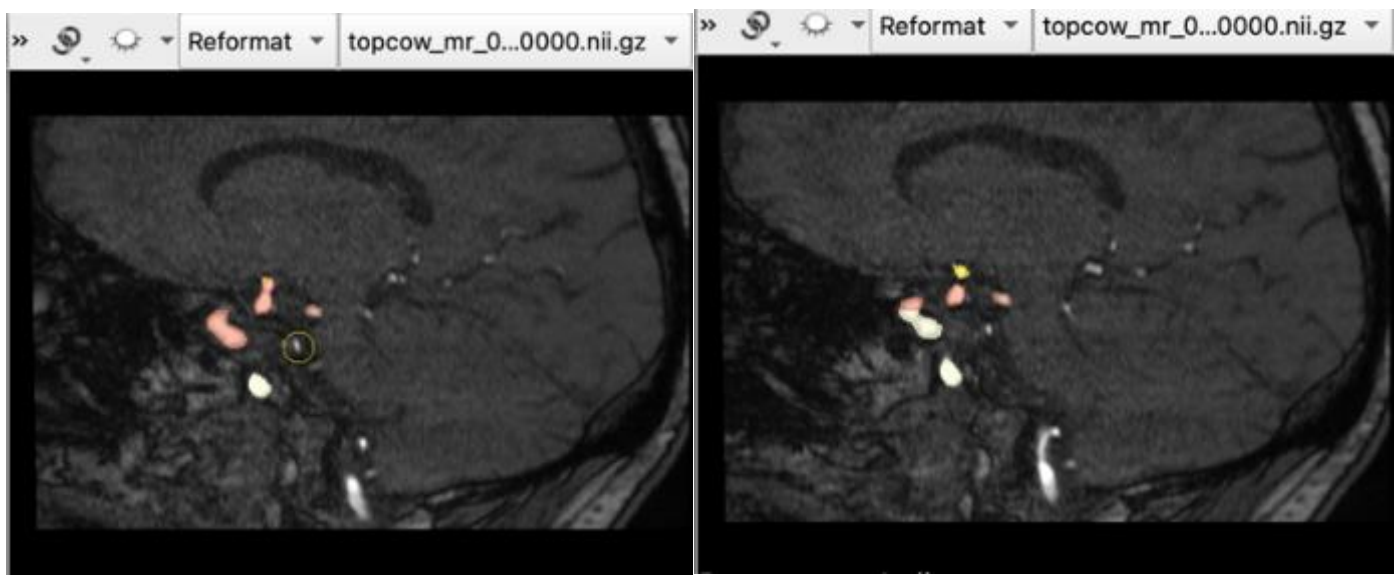

3. Extend MCAs to ~1cm past bifurcation.

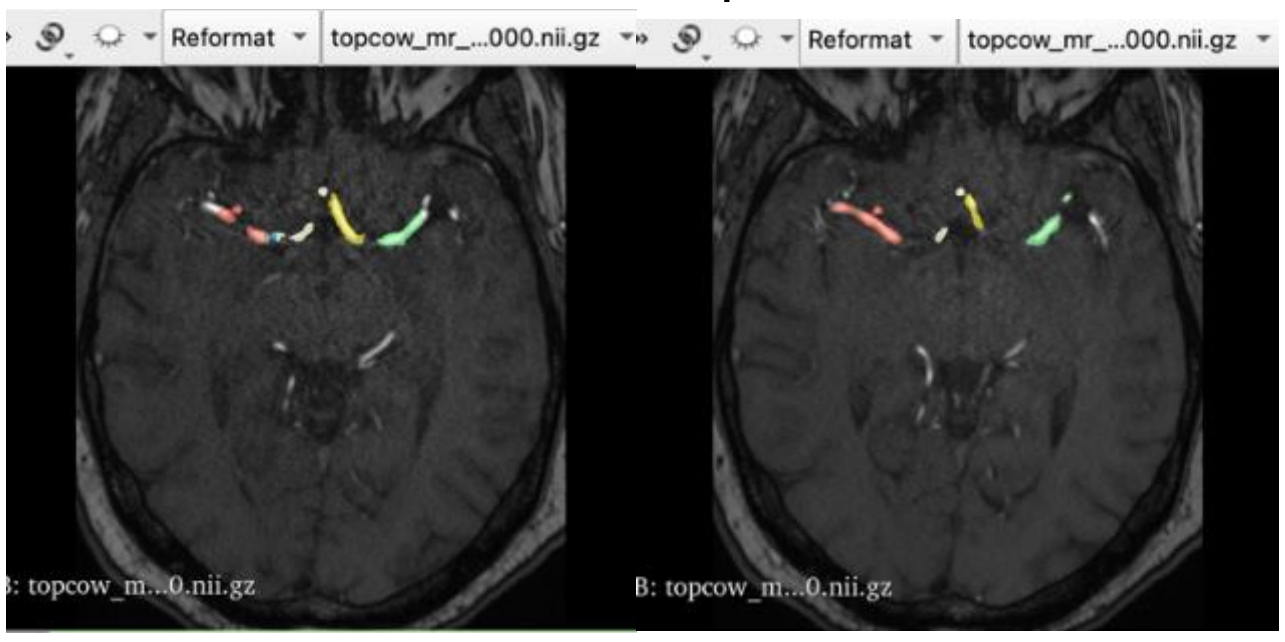

4. Extend "basilar artery" label to prox basilar, V4 verts, and prox 2-3 cm of large caliber SCAs, AICAs, and PICAs.

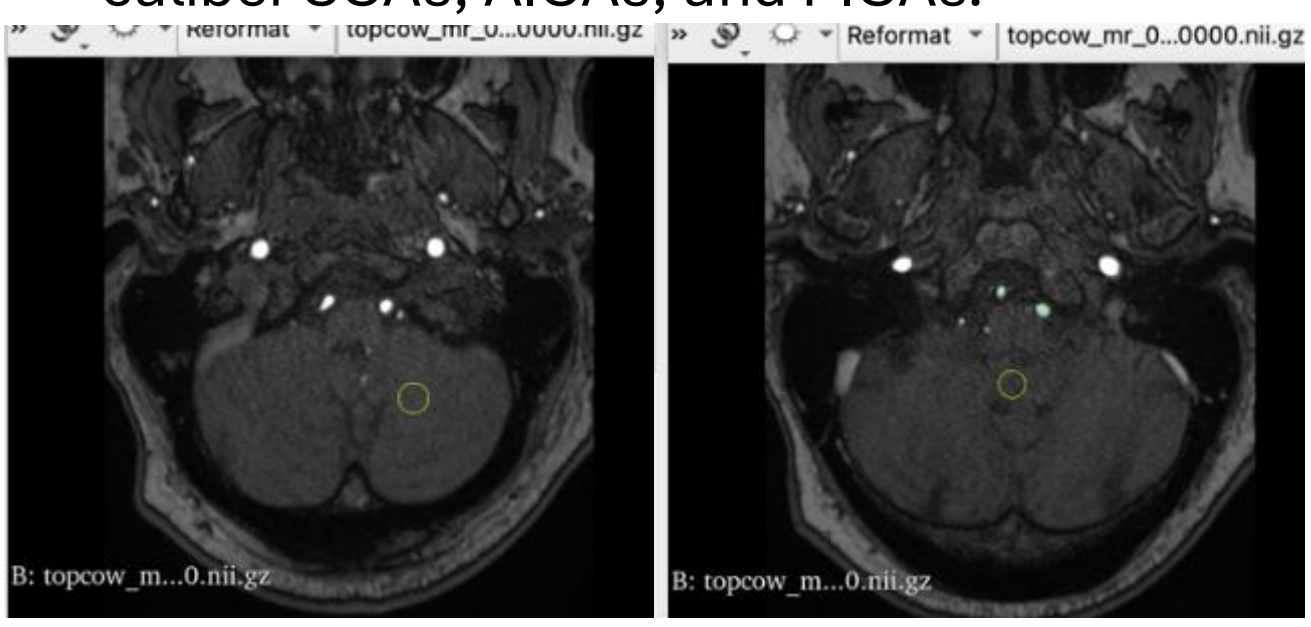

5. Add "basilar tip" label (from origin of SCAs up), which will require overwriting part of the basilar artery label.

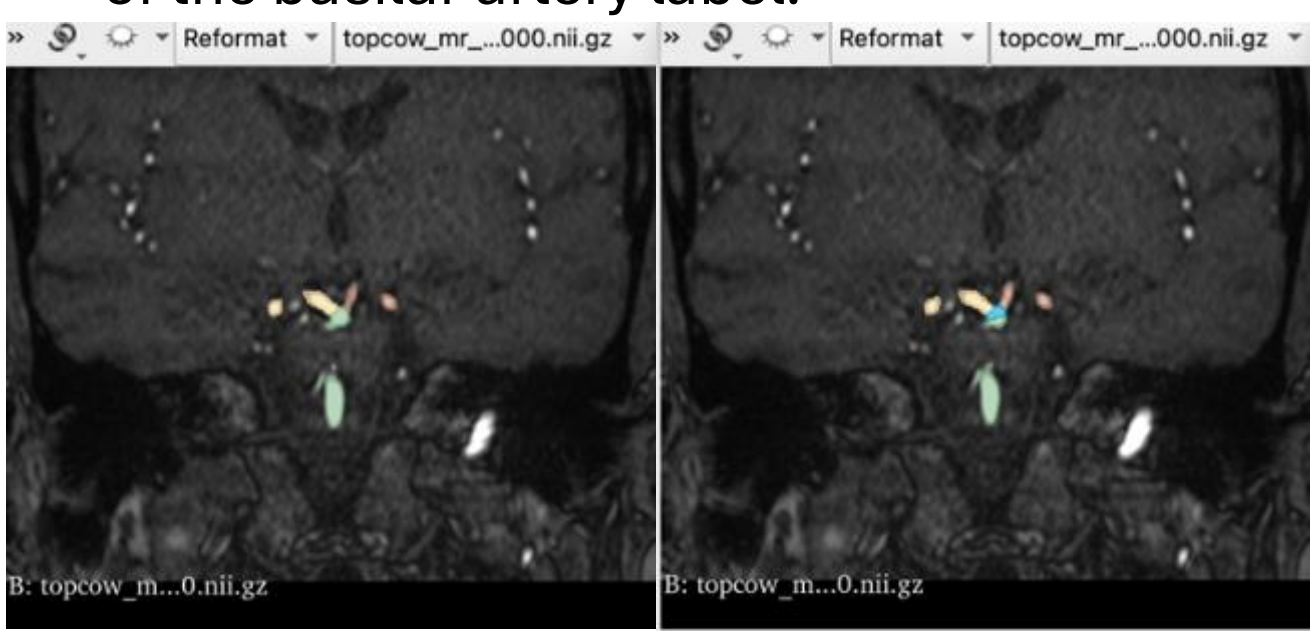

6. If there is a "3rd ACA" label in your dataset (small minority) either remove it or separate into right and left and add to respective ACA.